\documentclass[jimaging,article,accept,moreauthors,pdftex]{Definitions/mdpi} 

\usepackage{tikz}
\usepackage{mathtools}
\usepackage[ruled,vlined]{algorithm2e}
\usepackage{pgfplots}
\usepackage{pgfplotstable}
\pgfplotsset{compat=1.3}
\usepackage{booktabs}
\usepackage{float}
\usepackage{multirow}
\usepackage{graphicx}
\usepackage{authblk}

\usepackage{amsmath}
\usepackage{amsfonts}

\usetikzlibrary{calc,positioning}

\usepackage{xcolor}

\firstpage{1} 
\pubvolume{8}
\issuenum{3}
\articlenumber{64}
\pubyear{2022}
\copyrightyear{2022}
\externaleditor{Academic Editors: Jonathan Wu, Thangarajah Akilan, Jitendra Kumar and Chengsheng Yuan}
\datereceived{28 October 2021} 
\dateaccepted{27 February 2022} 
\datepublished{4 March 2022} 
\hreflink{https://doi.org/10.3390/\linebreak jimaging8030064} 

\Title{Rethinking Weight Decay for Efficient Neural Network Pruning}

\TitleCitation{Rethinking Weight Decay for Efficient Neural Network Pruning}

\Author{Hugo Tessier 
 $^{1,2,}$*{\orcidA{}}, Vincent Gripon $^{2}$\orcidB{}, Mathieu L\'{e}onardon $^{2}$\orcidC{}, Matthieu Arzel $^{2}$, Thomas Hannagan $^{1}$\orcidE{} \mbox{and David Bertrand $^{1}$}}

\AuthorNames{Hugo Tessier, Vincent Gripon, Mathieu L\'{e}onardon, Matthieu Arzel, Thomas Hannagan and David Bertrand}

\AuthorCitation{Tessier, H.; Gripon, V.; L\'{e}onardon, M.; Arzel, M.; Hannagan, T.; Bertrand, D.}

\address{%
	$^{1}$ \quad Stellantis, Centre Technique Vélizy, 78140 Vélizy-Villacoublay, France; \mbox{thomas.hannagan@stellantis.com (T.H.);} david.bertrand@stellantis.com (D.B.)\\
	$^{2}$ \quad IMT Atlantique, Lab-STICC, UMR CNRS 6285, 29238 Brest, France
; vincent.gripon@imt-atlantique.fr (V.G.); mathieu.leonardon@imt-atlantique.fr (M.L.); matthieu.arzel@imt-atlantique.fr (M.A.)}

\corres{Correspondence: hugo.tessier@imt-atlantique.fr; Tel.: +33-7494-00545}

\abstract{Introduced in the late {1980s} 
	 for generalization purposes, pruning has now become a staple for compressing deep neural networks. Despite many innovations in recent decades, pruning approaches still face core issues that hinder their performance or scalability. Drawing inspiration from early work in the field, and especially the use of weight decay to achieve sparsity, we introduce Selective Weight Decay (SWD), which carries out efficient, continuous pruning throughout training. Our approach, theoretically grounded on Lagrangian smoothing, is versatile and can be applied to multiple tasks, networks, and pruning structures. We show that SWD compares favorably to state-of-the-art approaches, in terms of {performance-to-parameters} ratio, on the CIFAR-10, Cora, and ImageNet ILSVRC2012 datasets.}


\begin{document}
	
	\section{Introduction}

	In recent decades, deep neural networks have become the reference for many machine learning tasks, especially computer vision. Their popularity quickly grew once deep convolutional networks managed to outclass classical methods on benchmark tasks, such as image classification on the ImageNet dataset \cite{krizhevsky2017imagenet}. Since their introduction by Le~Cun~et~al.~\cite{lecun1999object}, many architectural innovations have now contributed to their performance and efficiency~\mbox{\cite{he2016deep, ioffe2015batch, lin2013network, simonyan2014very, szegedy2015going, xie2017aggregated}.}
	However, for any given type of deep neural network architecture, the number of parameters tends to correlate with performance, resulting in the best-performing networks having prohibitive requirements in terms of memory footprint, computation power, and energy consumption~\cite{boukli2019processing}.

	This is a crucial issue for multiple reasons. Indeed, many applications, such as autonomous vehicles, require networks that can provide adequate, real-time responses on energy-efficient hardware: for such tasks, one cannot afford to have either an accurate network that is too slow to run or one that performs quickly but crudely. Additionally, research on deep learning relies heavily on iterative experiments that require a lot of computation time and power: lightening the networks would help to speed up the whole process.
	
	Many approaches have been proposed to tackle this issue. These include techniques such as distillation~\cite{hinton2015distilling, lassance2019deep}, quantization~\cite{courbariaux2016binaryconnect, rastegari2016xnornet}, factorization~\cite{denton2014exploiting}, and pruning \cite{han2015learning}; most of them can be combined~\cite{han2016deep}. The whole field tends to indicate that there may exist a Pareto optimum, between performance, memory occupation, and computation power, that compression could help to attain. However, progress in the field shows that this optimum has yet to be reached.
	
	Our work focuses on pruning. The basis of most pruning methods is to train a network and, according to a certain criterion, to identify which parts of it contribute the least to its performance. These parts are then removed and the network is fine-tuned to recover the incurred loss in performance~\cite{han2015learning, liu2017learning}. 
	
	Multiple decades of innovation in the field have uncovered many issues at stake when pruning networks, such as structure \cite{li2017pruning}, scalability \cite{liu2019rethinking, gale2019state}, or continuity \cite{louizos2018learning}. However, many approaches, while trying to tackle these issues, tend to resort to complex methods involving intrusive processes that make them harder to actually use, re-implement, and adapt to different networks, datasets, or tasks.
	
	Our contribution aims to solve these key problems in a more straightforward and efficient way that avoids human intervention in the training process as much as possible.
	Our method, Selective Weight Decay (SWD), is a pruning method for deep neural networks that is based on Lagrangian smoothing. It consists in a regularization which, at each step during the training process, penalizes the weights that would be pruned according to a given criterion. The penalization grows in the course of training until the magnitude of the targeted parameters is so close to zero that pruning them induces no drop in performance. This method has many desired properties, including avoiding any discontinuity, since pruned weights are progressively nullified.
	The weight removal is, itself, learned, which reduces the manual aspects of the pruning process. Moreover, since the penalized weights are not completely removed before the very end of training, the subset of the targeted parameters can {be adjusted} 
	 during training, depending on the current distribution of the weight magnitudes. The dependencies between weights can, thus, be better taken into~account.
	
	Our experiments show that SWD works well for both light-weight and large-scale datasets and networks with various pruning structures. Our method shines especially for aggressive pruning rates (few remaining parameter targets) and manages to achieve great results with targets for which classical methods experience a large drop in performance.
	
	Therefore, about SWD, which prunes deep neural networks continuously during training, we have the following claims:
	\begin{itemize}
		\item using standardized benchmark datasets, we prove that SWD performs significantly better on aggressive pruning targets than standard methods;
		\item we show that SWD needs fewer hyperparameters, introduces no discontinuity, needs no fine-tuning, and can be applied to any pruning structure with any pruning criterion.
	\end{itemize}
	
	In the following sections, we will review in detail the field of network pruning, describe our method, present our experiments and their results, and then discuss our observations.
	
	\section{Problem Statement and Related Work}\label{problemstatement}
	
	We now review the main pruning methods and attempt to organize them into \mbox{sub-families.}
	
	\subsection{Notations}
	
	We first recall the standard optimization problem with weight decay. Let $\mathcal{N}$ be a network with parameters $\mathbf{w}$, trained over dataset $\mathcal{D}$ containing $N$ pairs of input/groundtruth pairs $(x_i, y_i)$. The network is trained through error function $\mathcal{E}$, and penalized by a weight decay with a coefficient $\mu$ ~\cite{NIPS1991_563, Plaut1986ExperimentsOL}. The training process thus involves minimizing the objective function $\mathcal{L}$, defined as:
	\begin{equation}
		\mathcal{L}(\mathbf{w})=\underbrace{\sum_{(x,y)\in \mathcal{D}}\mathcal{E}(\mathcal{N}(x, \mathbf{w}), y)}_{\mathclap{\mathit{Err}\text{: error term}}} + \underbrace{\vphantom{\sum_{(x,y)\in \mathcal{D}}}\mu\left\|\mathbf{w}\right\|_2}_{\mathclap{\mathit{WD}\text{: weight decay}}}.
		\label{opti1}
	\end{equation} 
	
	\subsection{The Birth and Rebirth of Pruning}\label{birth}
	
	Although network size correlates with performance, the fundamental observation that motivates pruning is that not all of a trained network's parts seem to be useful. Unnecessary parts may be removed without penalizing performance.
	
	At the end of the 1980s and at the beginning of the 1990s, the field of pruning quickly expanded from a few seminal studies~\cite{NIPS1992_647, optimalbd, NIPS1988_119}. At the time, as observed by Reed~\cite{248452}, two major branches cohabited: (1) sensitivity calculation methods, which consisted in evaluating the contribution of each parameter to the error function and in pruning those which contributed the least, and (2) penalty-term methods, which penalized weights globally so as to encourage convergence to networks having a few big weights rather than lots of small ones. It is worth mentioning that pruning was originally intended to help the generalization of networks, rather than being a compression method per se.
	
	This field of research seems to have almost completely vanished during the ensuing decade, only to be resurrected by Han~et~al.~\cite{han2015learning}. Since then, the number of pruning investigations has expanded so quickly that it has made any reviewing task a challenging one~\cite{blalock2020state}.
	
	Assume we want to train and prune a given model with a target pruning rate $T$. The method of Han~et~al.~\cite{han2015learning}, which is currently the prototypical pruning technique, consists first in training, then in pruning and fine-tuning the network iteratively, with each time an increasing pruning rate $t$ until $T$ is reached. In particular, pruning is achieved here by reducing to and then maintaining at zero a proportion of the parameters of the whole network whose absolute magnitude is the smallest. 
	Though it is still possible to prune and fine-tune the model only once, doing so can be viewed as a particular case of the method.
	
	The literature that followed the work of Han~et~al.~\cite{han2015learning} has highlighted many questions that tend to be raised when pruning a neural network: ``Which parameters should be pruned?'', ``How can we prune them and  recover from the loss?'', and ``What kind of structures should be pruned?'' We will tackle these questions.
	
	\subsection{Which Parameters Should Be Pruned?}\label{criterion}
	
	One crucial prerequisite to pruning networks is to have a good criterion to define which parameters to prune. Many pruning criteria have been tested~\cite{anwar2016compact, hu2016network, luo2017thinet, srinivas2015data, yu2018nisp}, for example: Anwar~and~Sung~\cite{anwar2016compact} try various random masks and select the one which induces the least degradation; Hu~et~al.~\cite{hu2016network} prune on the basis of the average rate of null activation after each pruned layer. The two most widespread criteria are gradient magnitude and weight magnitude, both of which we will detail.
	
	\textls[-15]{The early branch of sensitivity calculation methods, birthed by the studies of Le~Cun~\cite{optimalbd} and Mozer~and~Smolensky~\cite{NIPS1988_119} and then studied within multiple articles~\cite{NIPS1992_647, NIPS1993_749, 80236,NIPS1996_1181}, led some recent studies to prune the weights of the least back-propagated gradient~\cite{dong2017learning, molchanov2017pruning}. Nevertheless, the criterion that remains the most common, namely, the mere magnitude of the parameters, turns out to be surprisingly effective while also intuitive. Although re-introduced by Han~et~al.~\cite{han2015learning}, it was first introduced by Chauvin~\cite{NIPS1988_133} and Hanson~and~Pratt~\cite{NIPS1988_156}, then presented under the name of {``clipping''} 
 by Janowsky~\cite{PhysRevA.39.6600}. Segee~and~Carter~\cite{155374} observed the surprising correlation between this intuitive criterion and that of Mozer~and~Smolensky~\cite{NIPS1988_119}, which is more theoretically grounded. These studies tend to confirm that magnitude is a good proxy for the contribution of a parameter to optimization problems summed up by Equation~\eqref{opti1}, which is why we used it in our experiments.}
	
	The other main branch identified by Reed~\cite{248452} revolved around enforcing sparsity using various kinds of weight decay regularization. The commonly stated motivation was that, if a certain parameter contributes poorly to the error term $\mathit{Err}$, then the weight decay term should outweigh it so that this very parameter would decrease toward zero.
	
	Since weight decay is required for weight-magnitude pruning, which is the favored criterion among several of the best implementations, it seems that sparsity-inducing regularizations are worth exploring further.
	
	\subsection{How to Prune Parameters and Recover from the Loss}\label{discontinuity} 
	
	One may object by noting that removing weights, even those that seem the least important, may damage the network in such a way that no fine-tuning could ever allow it to recover. Indeed, doing so severely disrupts the training process, for example, by removing parts of the network while it is trying to learn to solve a problem.
	The work of Le Cun, Bengio, and Hinton~\cite{lecun2015deeplearning} tends to show that the less the training process is disrupted, the better it~performs.
	
	For example, there is no guarantee that weights which seemed unimportant at first could not become crucial again in the new context of the pruned network. That is the reason why many efforts have focused on allowing weights to regrow in one way or another. Different approaches have been proposed~\cite{bellec2018deep, dai2018nest, he2018soft} to either regrow previously pruned weights or to not completely prune parameters by still allowing them to be trained once they are reduced to zero by pruning.
	
	The principle of regrowing weights is central to the family of methods that could be called \textit{sparse training}. \textit{Sparse training} was first introduced by Mocanu~et~al.~\cite{mocanu} and then further explored within the literature~\cite{dettmers2019sparse, evci2020rigging, mostafa2019parameter}. It involves training the network with a constant level of sparsity, at first spread randomly with uniform probability, and then adjusted during steps which combine (1) pruning of a certain portion of the weights, according to a certain criterion, and (2) regrowing an equivalent amount of weights, depending on another~criterion.
	
	Such a family of methods provided a promising way to work around the problem of falsely unimportant weights, while limiting the impact of increasing sparsity all throughout~training.
	
	Of course, sparse training is not the only family of methods to tackle this issue. Indeed, this technique belongs to a vast trend in the literature: discovering the importance of sparsity during training to achieve better results with shrunken networks.
	Pruning networks early, so that they start training with their definitive sparsity from the beginning, is the whole point of a whole range of work~\cite{frankle2020pruning, lee2019snip, tanaka2020pruning, wang2019picking}, as well as one of the motivations behind the field surrounding the \textit{lottery ticket hypothesis}~\cite{frankle2018lottery, frankle2020linear, frankle2019stabilizing, malach2020proving, morcos2019one, zhou2019deconstructing}. Renda~et~al.~\cite{renda2020comparing}, while studying the lottery ticket hypothesis, came up with a method called \textit{learning rate rewinding}, which proposes replacing the fine-tuning step with a full retraining stage that uses the weights of the trained and pruned network as a new initialization.
	
	Another distinct branch of methods involves work that aims to induce sparsity in a more continuous way throughout the training process, possibly avoiding any fine-tuning or retraining. One intuition that motivates these methods is that delegating the care of sparsity to the gradient descent is a sensible way to not disrupt training too much.
	
	One sub-family of this branch focuses on finding a way to learn a pruning mask during training \cite{guo2016dynamic, louizos2018learning, srinivas2016training, NIPS2019_9521}; some of these studies propose learning such a mask using variants of the quantification method of Courbariaux~et~al.~\cite{courbariaux2016binaryconnect} on auxiliary learnable parameters.
	The other major sub-family counts methods grounded on a Bayesian mathematical formalism~\cite{dai2018compressing, louizos2017bayesian, molchanov2017variational, neklyudov2017structured,ullrich2017soft}. They mainly consist in various kinds of sparsity-inducing regularization, whose parameters are tuned through variational inference. These methods are the ones that stick most closely to the former family of penalty-term methods: Ullrich~et~al.~\cite{ullrich2017soft} even references the work of Nowlan and Hinton~\cite{nowlan1992simplifying}. 
	
	However, this whole family of methods tends not to be among the simplest to implement and adapt to various kinds of tasks, datasets, or structures. Adapting variational dropout~\cite{molchanov2017variational} to structured pruning is the focus of whole contributions~\cite{louizos2017bayesian, neklyudov2017structured}, and Gale~et~al.~\cite{gale2019state} show that the work of Molchanov~et~al.~\cite{molchanov2017variational} and Louizos~et~al.~\cite{louizos2018learning} does not scale easily to large datasets such as ImageNet ILSVRC2012.
	
	Unfortunately, one problem that encompasses the whole field, and about which Blalock~et~al.~\cite{blalock2020state} raise the alarm, is the lack of comparability between the various methods in the literature.
	Indeed, contributions to the domain rarely compare to the same reference methods, tasks, models, training conditions, or datasets and do not always show the same metrics computed in a consistent way. Hence, it is very difficult to know which method brings actual methodological or theoretical improvements on the topic, and it appears that none of the questions or aspects we mentioned can be considered solved for now. Therefore, while taking inspiration from these methods and their desirable properties, we propose one that is easier to scale, adapt, and apply.
	
	\subsection{What Kind of Structures Should Be Pruned?}\label{granularity}
	
	As pointed out by both Anwar~et~al.~\cite{anwar2015structured} and Li~et~al.~\cite{li2017pruning}, the parameter-wise pruning of Han~et~al.~\cite{han2015learning} produces sparse matrices that are hardly exploitable by modern hardware and deep learning frameworks. That is why a whole field of pruning is dedicated to finding ways to eliminate parts of the networks in a structured way that can actually induce a measurable speedup.
	
	Because of the predominance of convolutional neural networks in the literature, the most widespread type of structure to be considered by the field is convolutional channels (or \textit{filters})~\cite{anwar2015structured, anwar2016compact, he2018soft,he2017channel,huang2018learning,li2017pruning, liu2017learning, luo2017thinet, yamamoto2019pcas}. Indeed, pruning filters induces a direct shrinking of the very architecture of the network and a quadratic reduction in the parameter count, as each removed filter is one less input feature map for the next convolution layer.
	
	Other types of structures have been experimented on, such as kernels or intra-kernel strided structures~\cite{anwar2015structured,anwar2016compact} or the reduction of convolutions to shift operations~\cite{hacene2019attention}. In this work, we focus on two granularity levels: parameter-wise (\textit{unstructured}) and convolutional filter-wise (\textit{structured}).
	
	\section{Selective Weight Decay}\label{swd}
	
\begin{figure}[H]
	\begin{tikzpicture}[scale=3]
		\draw[->] (-1, -1) -- (1, -1) node[right] {$w$};
		\draw[->] (0, -0.95) -- (0, 0.5) node[above] {$p$};
		
		\draw[->, dashed, red] (0.4, -0.27) -- (0.4, -0.37) node[above] {};
		\draw[->, dashed, red] (-0.4, -0.27) -- (-0.4, -0.37) node[above] {};
		
		\draw[->, dashed, red] (0.25, -0.5625) -- (0.25, -0.7125) node[above] {};
		\draw[->, dashed, red] (-0.25, -0.5625) -- (-0.25, -0.7125) node[above] {};		
		
		\draw[->, dashed, red] (0.1, -0.72) -- (0.1, -0.92) node[above] {};
		\draw[->, dashed, red] (-0.1, -0.72) -- (-0.1, -0.92) node[above] {};	
		
		\draw[scale=1, domain=-0.5:0.5, smooth, very thin, dash pattern=on 1pt off 4pt,variable=\x, red] plot ({\x}, {0.5*\x*\x - 0.5*0.5*0.5});
		\draw[scale=1, domain=-0.5:0.5, smooth, thin, dash pattern=on 1pt off 3pt, variable=\x, red] plot ({\x}, {1*\x*\x - 1*0.5*0.5});
		\draw[scale=1, domain=-0.5:0.5, smooth, thick, dash pattern=on 1pt off 2pt, variable=\x, red] plot ({\x}, {2*\x*\x - 0.5});
		\draw[scale=1, domain=-0.5:0.5, smooth, thick, variable=\x, red] plot ({\x}, {3*\x*\x - 3*0.5*0.5});

		\draw[scale=1, domain=-0.5:0.5, smooth,  thin, dash pattern=on 1pt off 4pt, variable=\x, green] plot ({\x}, {0.3*\x*\x -0.075});
		\draw[scale=1, domain=0.5:1, smooth, thick, variable=\x, green] plot ({\x}, {0.3*\x*\x -0.075});
		\draw[scale=1, domain=-1:-0.5, smooth, thick, variable=\x, green] plot ({\x}, {0.3*\x*\x -0.075});
		
		\draw[scale=1, domain=-0.95:0.5, smooth, dashed, variable=\y, black] plot ({0.5}, {\y});
		\draw[scale=1, domain=-0.95:0.5, smooth, dashed, variable=\y, black] plot ({-0.5}, {\y});
		
		\node[text width=0.5cm, green] at (0.8,0.22) {\small WD};
		\node[text width=0.5cm, red] at (0.37,-0.6) {\small SWD};
		
		\node[text width=0.5cm] at (0.62,-0.95) {$t$};
		\node[text width=0.5cm] at (-0.60,-0.95) {$-t$};
		
	\end{tikzpicture}
		\caption{To prune deep neural networks continuously during training, we apply distinct types of weight decay (penalty $p$ on the y-axis) depending on weight magnitude (weight value $w$ on the x-axis). Weights whose magnitude exceeds a threshold $t$ (defined according to the number of weights to prune) are penalized by a regular weight decay. Those beneath this threshold are targeted by a stronger weight decay whose intensity grows during training. This stronger weight decay, only applied to a subset of the network, is the Selective Weight Decay. This approach can be equally well applied to weights (unstructured case) or groups of weights (structured case).}
		\label{introFigure}
	\end{figure}
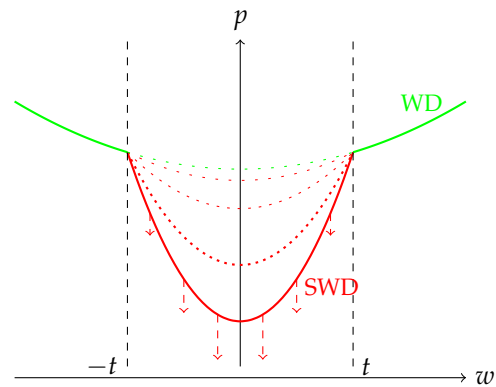

	We now present our contribution, illustrated in Figure~\ref{introFigure}, and explain how it addresses the aforementioned issues.
	
	\subsection{Principle}
	
	Selective Weight Decay (SWD) is a regularization which induces sparsity continuously on any type of structure: at each training step, a certain penalization is applied to the parameters to be pruned at this very step according to a certain criterion and a certain structure. The criterion we chose is weight magnitude, or variants of it according to the chosen structure. The penalized optimization problem can be viewed as:
	\begin{equation}
		\mathcal{L}(\mathbf{w})=\underbrace{\sum_{(x,y)\in \mathcal{D}}\mathcal{E}(\mathcal{N}(x, \mathbf{w}), y)}_{\mathclap{\mathit{Err}}}
		+ \underbrace{\vphantom{\sum_{(x,y)\in \mathcal{D}}}\mu\left\|\mathbf{w}\right\|_2}_{\mathclap{\mathit{WD}}}
		+ \underbrace{\vphantom{\sum_{(x,y)\in \mathcal{D}}}a\mu\left\|\mathbf{w}^*\right\|_2}_{\mathclap{\mathit{SWD}}},
		\label{opti2}
	\end{equation}
	with $a$ being a coefficient which determines the importance of SWD relative to the rest of the optimization problem. $\mathbf{w}^*$ is the subset of $\mathbf{w}$ to be pruned at a certain step. SWD is summed up as Algorithm~\ref{algo3}.
	\begin{algorithm}[h]
		\KwData{network $\mathcal{N}$ of weights $\mathbf{w}$, dataset $\mathcal{D}$, target pruning rate $T$}
		$a \gets a_{min}$\; 
		\SetAlgoLined
		\While{the network is not fully trained and $a \leq a_{max}$}{
			draw batches $x$ and $y$ from $\mathcal{D}$\; 
			$\mathit{Err} \gets \mathcal{E}(\mathcal{N}(x, \mathbf{w}), y) $\; 
			$\mathit{WD} \gets \mu\left\|\mathbf{w}\right\|_2$\; 
			determine $\mathbf{w}^*$ according to $T$\; 
			$\mathit{SWD} \gets \mu\left\|\mathbf{w}^*\right\|_2$\; 
			backpropagate $\mathit{Err} + \mathit{WD} + a\mathit{SWD}$\; 
			update weights\; 
			increase $a$\; 
		}
		\caption{Summary of SWD}
		\label{algo3}
	\end{algorithm}

	
	The evolution of $a$ is designed to be exponential and, according to two bounds $a_{min}$ and $a_{max}$ at a certain training step $s$, is defined as such:
	\begin{equation}
		a(s) = a_{min} \left( \frac{a_{max}}{a_{min}}\right)^\frac{s}{s_{final}},
		\label{progressive}
	\end{equation}
	with $s_{final}$ being the value at which SWD reaches $a_{max}$ and, usually, the total number of training iterations.
	The exponential increase in \mbox{Equation \eqref{progressive}} allows the network to converge before applying a strong penalization. We favored an exponential increase over a linear one so that $a$ can reach large final values without penalizing  training too much throughout the training process. In addition, setting $a_{min}$ and $a_{max}$ manually allows precise control over the evolution of the penalty and enables a careful study of how training behaves under this~constraint.
	
	\subsection{SWD as a Lagrangian Smoothing of Pruning}
	
	Penalizing weights until they reach zero appears to be a viable method to relax the hard constraint that is pruning. Indeed, pruning can be seen as a constraint on the $\mathcal{L}_0$ norm of the network, which is non-differentiable (this is a problem in differentiable optimization methods such as those used in deep learning). The $\mathcal{L}_2$ norm can be used as a differentiable relaxation of $\mathcal{L}_0$. We designed SWD so that it can be viewed as a Lagrangian smoothing, with coefficient $a$ of the SWD term in Equation~\eqref{opti2} being a Lagrangian multiplier.
	
	As pointed out by the work of Murray~and~Ng~\cite{lagrangian}, Lagrangian smoothing allows convergence relative to both the error term and the constraint. While $a$ can be mostly negligible at the start of the training, it becomes preponderant at the end and forces the target weights to be pruned, so as not to hinder the convergence of the network during training while allowing for pruning.
	The fact that SWD only penalizes weights selectively during training allows them to recover as soon as they are no longer targeted by the pruning criterion, thereby combining both the pruning and regrowing criteria of sparse training. Therefore, SWD is a non-greedy method that allows weights to recover when needed.
	
	\subsection{On the Adaptability of SWD to Structures}\label{SWDstructure}
	
	The exact definition of $\mathbf{w}^*$ depends on the chosen type of structure to prune. As SWD induces no constraint on such considerations, it can be applied to any type of structure.
	
	Unstructured pruning is defined by removing all the weights of least magnitude in the whole network so that the proportion of pruned parameters matches the pruning target as closely as possible. Formally stated:
	\begin{equation}
		\begin{gathered}
			\mathbf{w}^*=\left\{|w| \leq \delta, \; w \in \mathbf{w}\right\}, \\
			\text{with} \; \delta \; \text{so that} \; n(\mathbf{w}^*)=T n(\mathbf{w}),
		\end{gathered}
	\end{equation}
	with $T$ being the pruning target and $n(\mathbf{w})$ the number of elements of the parameters $\mathbf{w}$.
	
	We based the structured version of our SWD on the method of Liu~et~al.~\cite{liu2017learning} to solve a problem induced by residual connections in modern convolutional networks: to ensure that the exact output dimensions of the feature maps after each residual connection match the desired target, one must prune exactly the same channels among the connections and the last convolution before them. To the best of our knowledge, this problem has not been tackled, and approaches that use certain norms of the filters as a criterion could not be adjusted to tackle this important problem without altering them too drastically, if the desire was to remain true to the original contribution.
	
	However, since the method of Liu~et~al.~\cite{liu2017learning} prunes multiplicative coefficients of batchnorm layers, it is easy for it to solve the residual connection issue as soon as a batchnorm layer is inserted after each residual connection (which does not change the overall performance of the network).
	
	Han~et~al.~\cite{liu2017learning} considered the magnitudes of multiplicative coefficients in a batchnorm layer to be an estimator of the importance of their corresponding filters. These batchnorm layers were then penalized during training by a smooth-$L_1$ norm. In their work, a global threshold was applied to all the batchnorm layers in order to globally prune  a target percentage of all the smallest batchnorm coefficients.
	
	However, in order to have  better control over the exact number of parameters at the end of the pruning process, we instead prune all the smallest batchnorm layers until a portion of the overall network (once the parameters of the corresponding convolutional filters have been substracted) is removed.
	
	
	\section{Experiments}\label{experiments}
	
	\subsection{General Training Conditions}\label{conditions}
	
	In order to eliminate all unwanted variables, each series of experiments was run under the same conditions, except when explicitly stated, with the same initialization and same seed for the random number generators of the various used libraries. We used no pre-trained networks and we trained all of them in a very standard way.
	
	The training conditions were as follows: all our networks were trained using the Pytorch framework (Paske~et~al.~\cite{paszke2019pytorch}); using SGD as an optimizer, with a base learning rate of {$1\times 10^{-1}$} for the first third of the training, then {$1\times 10^{-2}$} for the second, and finally \mbox{{$1\times 10^{-3}$}} for the last third, and momentum set to 0.9. All networks are initialized using default initialization from Pytorch. {Our code is available at} \url{https://github.com/HugoTessier-lab/SWD}{, accessed on 26 February 2022}. 

	\subsection{Chosen Baselines and Specificities of Each Method}\label{spec}
	
	\paragraph{\textbf{Unstructured pruning: Han~et~al. 
~\cite{han2015learning}}} The networks trained with this method were pruned and fine-tuned for five iterations. At each step, the pruning target is a fraction of the final one: for example, when first pruned, only a fifth of the final pruning target is actually removed. The pruned weights are those of least magnitude.
	\vspace{-5pt}
	\paragraph{\textbf{Structured pruning: Liu~et~al.~\cite{liu2017learning}}} The network is only pruned once and fine-tuned. In accordance with the paper, a smooth-$L_1$ norm is applied as a regularization to every prunable batchnorm layer with a coefficient $\lambda$ that depends on the dataset. This method appeared to us to be the most straightforward one for allowing an accurate evaluation of the number of pruned parameters. The pruned filters are those whose multiplicative coefficients in the batch normalization layer are of least magnitude.
	\vspace{-5pt}
	\paragraph{\textbf{LR-Rewinding: Renda~et~al.~\cite{renda2020comparing}}} When networks are trained following this method, the post-removal fine-tuning is replaced by a retraining which consists in repeating  the pre-removal training process exactly, with the same learning rate values and the same number of epochs. This method updates and significantly improves the train, prune, and fine-tune framework that serves as a basis for both previous methods.
	\vspace{-5pt}
	\paragraph{\textbf{SWD}} 
	Whether on unstructured or structured pruning, 
	when trained with SWD, the network is not fine-tuned at all and only pruned once at the end.
	The values $a_{min}$ and $a_{max}$ vary according to the model and the dataset.
	\vspace{-3pt}
	\paragraph{\textbf{Overall methodology}}
	
	In order to isolate the respective gain of each method:
		\begin{itemize}
			\item All the unstructured pruning methods use weight magnitude as their criterion;
			\item All the structured pruning methods are applied to batch normalization layers;
			\item Structured LR-Rewinding also applies the smooth-$L_1$ penalty from Liu~et~al.~\cite{liu2017learning};
			\item The hyper-parameters specific to the aforementioned methods, namely, the number of iterations and the values of the smooth-$L_1$ norm, are directly extracted from their respective original papers.
	\end{itemize}
	
	Here are the only notable differences:
		\begin{itemize}
			\item SWD does not apply any fine-tuning;
			\item Unstructured LR-Rewinding only re-trains the network once (because of the extra cost from fully retraining networks, compared to fine-tuning);
			\item SWD does not apply a smooth-$\mathcal{L}_1$ norm (since it would clash with SWD's own~penalty).
		\end{itemize}

	\subsection{Comparison with the State of the Art}

	Table~\ref{totallyLegitTable} shows results from different techniques, as presented in the related papers, on different datasets, networks, compression rates, and pruning structures. To achieve the best performance possible, results of SWD in the case of structured pruning on ImageNet are ran with warm-restart and 180 epochs in total.
	
	\begin{table}[H]
		\caption{Quick comparison between SWD and multiple pruning methods, for different datasets and networks. All lines marked with an \textbf{*} are results obtained with our own implementations; all the others are extracted from the original papers.}
		\label{totallyLegitTable}
\small

\begin{tabular}{m{3cm}<{\centering}m{1.7cm}<{\centering}cccc}
	\toprule
	
	\textbf{Method} & \textbf{Type} & \textbf{Dataset} & \textbf{Network} & \textbf{Comp. }& \textbf{Accuracy} \\
	
	\midrule
	
	Liu et al.~\cite{liu2018frequency} & Unstructured & ImageNet & AlexNet & $\times$22.6 & 56.82\% (+0.24\%) \\
	Zhu et al.~\cite{zhu2017prune} & Unstructured & ImageNet & InceptionV3 & $\times$8 & 74.6\% ($-$3.5\%) \\
	Zhu et al.~\cite{zhu2017prune} & Unstructured & ImageNet & MobileNet & $\times$10 & 61.8\% ($-$8.8\%) \\

	Xiao et al.~\cite{NIPS2019_9521} & Unstructured & ImageNet & ResNet50 & $\times$2.2 & 74.50\% ($-$0.40\%) \\
	SWD (ours) \textbf{*}& Unstructured & ImageNet & ResNet50 & $\times$2 & 75.0\% ($-$0.7\%)\\
	SWD (ours) \textbf{*}& Unstructured & ImageNet & ResNet50 & $\times$10 & 73.1\% ($-$1.8\%)\\
	SWD (ours) \textbf{*}& Unstructured & ImageNet & ResNet50 & $\times$40 & 67.8\% ($-$7.1\%)\\
\midrule

	Liu et al.~\cite{liu2017learning} \textbf{*} & Structured & ImageNet & ResNet50 & $\times$2 & 63.6\% ($-$12.1\%) \\

	Luo et al.~\cite{luo2017thinet} & Structured & ImageNet & ResNet50 & $\times$ 2.06 & 72.03\% ($-$3.27\%)\\
	Luo et al.~\cite{luo2017thinet} & Structured & ImageNet & ResNet50 & $\times$ 2.95 & 68.17\% ($-$7.13\%)\\
	Molchanov et al.~\cite{molchanov2019importance} & Structured & ImageNet & ResNet50 & $\times$1.59 & 74.5\% ($-$1.68\%)\\
	Molchanov et al.~\cite{molchanov2019importance} & Structured & ImageNet & ResNet50 & $\times$2.86 & 71.69\% ($-$4.49\%)\\
	SWD (ours) \textbf{*} & Structured & ImageNet & ResNet50 & $\times$1.33 & 74.7\% ($-$1.0\%)\\
	SWD (ours) \textbf{*} & Structured & ImageNet & ResNet50 & $\times$2 & 73.9\% ($-$1.8\%)\\

	\bottomrule
\end{tabular}
	\end{table}
	\begin{table}[H]\ContinuedFloat
		\caption{{\em Cont.}}
		\label{totallyLegitTable}
\small

\begin{tabular}{m{3cm}<{\centering}m{1.7cm}<{\centering}cccc}
	\toprule
	\textbf{Method} & \textbf{Type} & \textbf{Dataset} & \textbf{Network} & \textbf{Comp. }& \textbf{Accuracy} \\
	
	\midrule
		Liu et al.~\cite{liu2017learning} & Structured & CIFAR10 & DenseNet40 & $\times$2.87 & 94.35\% (+0.46\%)\\
	Liu et al.~\cite{liu2017learning} & Structured & CIFAR10 & ResNet164 & $\times$1.54 & 94.73\% (+0.15\%)\\

	Ye et al.~\cite{ye2018rethinking} & Structured & CIFAR10 & ResNet20$-$16 & $\times$1.6 & 90.9\% ($-$1.1\%)\\
	Ye et al.~\cite{ye2018rethinking} & Structured & CIFAR10 & ResNet20$-$16 & $\times$3.1 & 88.8\% ($-$3.2\%)\\
	SWD (ours) \textbf{*} & Structured & CIFAR10 & ResNet20$-$16 & $\times$1.42 & 91.22\% ($-$1.15\%)\\ 
	SWD (ours) \textbf{*} & Structured & CIFAR10 & ResNet20$-$16 & $\times$3.33 & 88.93\% ($-$3.44\%)\\

	Liu et al.~\cite{liu2017learning} \textbf{*} & Structured & CIFAR10 & ResNet20$-$64 & $\times$2 & 94.92\% ($-$0.75\%)\\
	SWD (ours) \textbf{*} & Structured & CIFAR10 & ResNet20$-$64 & $\times$2 & 94.96\% ($-$0.71\%)\\ 
	SWD (ours) \textbf{*} & Structured & CIFAR10 & ResNet20$-$64 & $\times$50 & 89.07\% ($-$6.5\%)\\

	\bottomrule
\end{tabular}
	\end{table}

	Since these results do not come from identical networks on the same datasets, trained in the same conditions, and pruned at the same rate, the comparisons have to be interpreted with caution. However, Table~\ref{totallyLegitTable} gives quantified indications as to how our method compares to the state of the art, in terms of performance and allowed compression rates.
	
	\subsection{Experiments on ImageNet ILSVRC2012}
		
	The results of the experiments on ImageNet ILSVRC2012 are shown in Table~\ref{resnetBigResults}, which presents the top-one and top-five accuracies of ResNet-50 from He~et~al.~\cite{he2016deep} 
	on ImageNet ILSVRC2012, under the conditions described in Sections~\ref{conditions} and \ref{spec}.
	The ``Baseline'' method is a regular, non-pruned network, which serves as a reference. SWD outperforms the reference method for both unstructured and structured pruning.
	
	\begin{table}[H]
	
		\caption{Results 
 with ResNet-50 on ImageNet ILSVRC2012, with unstructured and structured pruning for different rates of remaining parameters. SWD outperforms the reference method (or its counterpart with additional LR-Rewinding) in both cases. All values are in \%. {The best performance for each target is indicated in bold.}}
		\label{resnetBigResults}

\begin{tabular}{m{2cm}<{\centering}m{2cm}<{\centering} cm{2.1cm}<{\centering} cm{2.1cm}<{\centering}c}
	\toprule
	\multicolumn{7}{c}{\textbf{Experiments on ImageNet ILSVRC2012}}\\
	
	\midrule
	\multicolumn{7}{c}{\textbf{Unstructured pruning}}\\
	
	\multirow{2}{*}{Target (\%)} & \multicolumn{2}{c}{Han et al.~\cite{han2015learning}}& \multicolumn{2}{c}{+LRR \cite{renda2020comparing}} & \multicolumn{2}{c}{SWD (ours)}\\
	\cmidrule{2-7}
	& Top-1& Top-5 & Top-1& Top-5 & Top-1& Top-5\\
	\midrule
	50 &74.9&92.2& 58.4&82.1&  \textbf{75.0}  &    \textbf{92.2}\\
	10 &71.1&90.5& 54.6&79.6&    \textbf{73.1}&     \textbf{91.3}\\
	2.5 &47.2&73.2& 34.8&61.54&     \textbf{67.8}&       \textbf{88.4}\\
	
	\midrule
	\multicolumn{7}{c}{\textbf{Structured pruning}}\\
	
	\multirow{2}{*}{Target (\%)} & \multicolumn{2}{c}{Liu et al.~\cite{liu2017learning}}& \multicolumn{2}{c}{+LRR \cite{renda2020comparing}} & \multicolumn{2}{c}{SWD (ours)}\\
	\cmidrule{2-7}
	& Top-1& Top-5 & Top-1& Top-5 & Top-1& Top-5\\
	\midrule
	90 &\textbf{74.7}&\textbf{92.2}& 56.1&80.7&  74.2  &    91.9\\
	75 &73.4&91.6& 51.1&77.1&    \textbf{73.5}&     \textbf{91.5}\\
	50 &63.6&85.7& 40.0&66.2&     \textbf{69.0}&       \textbf{88.8}\\
	20 &0.1&0.5& 0.1&0.5&     \textbf{69.0}&       \textbf{88.7}\\

%
%
%
%
%
%
%
%
%
%
	\bottomrule
\end{tabular}
	\end{table}

	For models trained on ImageNet ILSVRC2012, the standard weight decay (parameter $\mu$) is set to   {$1\times 10^{-4}$}
 and models were trained during {90} epochs.
	For the method from Han~et~al.~\cite{han2015learning}, we made each fine-tuning step last for 5 epochs, except for the last iteration which lasted 15 epochs. For Liu~et~al.'s~\cite{liu2017learning} method, the network was only pruned once and fine-tuned over 40 epochs. The smooth-$L_1$ norm had a coefficient set to {$\lambda=1\times 10^{-5}$}.
	For unstructured pruning, SWD was applied with {$a_{min}=1\times 10^{-1}$} and {$a_{max}=1\times 10^{5}$}; for structured pruning, {$a_{min}=10$} and {$a_{max}=1\times 10^{4}$}.

	\subsection{Impact of SWD on the Pruning/Accuracy Trade-Off}\label{cifarConditions}
		
	Even though obtaining better accuracy for a given pruning target is not without interest, it makes more sense to know what maximal compression rate SWD would allow for a given accuracy target. 
	
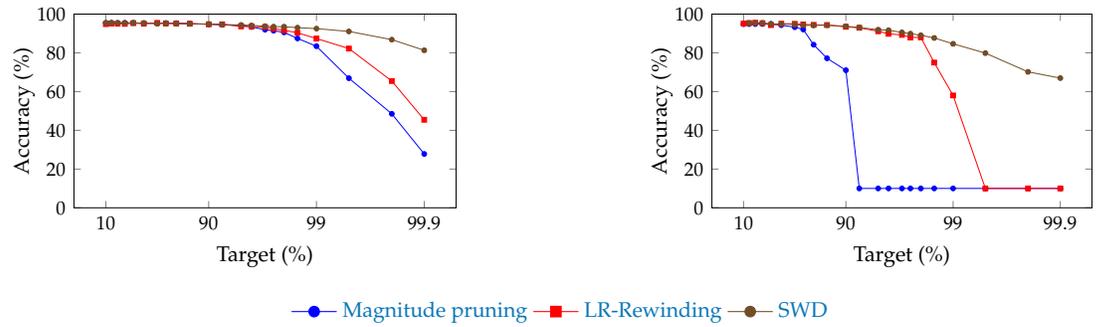
\begin{figure}[H]
			\pgfplotsset{footnotesize,samples=10}
\begin{subfigure}
	\centering
	\begin{tikzpicture}
		\begin{scope}[scale=0.9]
			\begin{axis}[
				xlabel=Target (\%),
				xmode=log,
				ymin=0,
				ymax=100,
				mark size = 1pt,
				xtick={90, 10, 1, 0.1},
				xticklabels={10,90,99,99.9},
				x dir=reverse,
				ylabel=Accuracy (\%),
				ylabel shift = -6 pt,
				width=0.5\textwidth,
				height=1.75in,
				legend entries={Magnitude pruning, LR-Rewinding, SWD},
				legend to name=named,
				legend columns=-1,
				legend style = {draw=none},
				]
				
				\addplot coordinates
				{
					(90 , 95.45)
					(80 , 95.47)
					(70 , 95.43)
					(60 , 95.48)
					(50 , 95.44)
					(40 , 95.32)
					(30 , 95.3)
					(25 , 95.32)
					(20 , 95.32)
					(15 , 95.05)
					(10 , 94.77)
					(7.5 , 94.48)
					(5 , 94.03)
					(4 , 93.38)
					(3 , 91.95)
					(2.5 , 91.43)
					(2 , 90.58)
					(1.5 , 87.44)
					(1 , 83.42)
					(0.5 , 66.9)
					(0.2 , 48.52)
					(0.1 , 27.78)
				};
				\addplot coordinates
				{
					(90 , 94.82)
					(80 , 95.15)
					(70 , 95.03)
					(60 , 94.94)
					(50 , 95.33)
					(40 , 95.04)
					(30 , 95.45)
					(25 , 95.15)
					(20 , 95.14)
					(15 , 95.03)
					(10 , 94.72)
					(7.5 , 94.74)
					(5 , 93.66)
					(4 , 93.63)
					(3 , 93.34)
					(2.5 , 92.48)
					(2 , 91.64)
					(1.5 , 90.36)
					(1 , 87.38)
					(0.5 , 82.21)
					(0.2 , 65.46)
					(0.1 , 45.44)
				};
				\addplot coordinates
				{(90 , 95.43)
					(80 , 95.55)
					(70 , 95.47)
					(60 , 95.40)
					(50 , 95.46)
					(40 , 95.37)
					(30 , 95.04)
					(25 , 95.34)
					(20 , 95.09)
					(15 , 94.99)
					(10 , 94.90)
					(7.5 , 94.58)
					(5 , 94.4)
					(4 , 94.14)
					(3 , 93.76)
					(2.5 , 93.52)
					(2 , 93.49)
					(1.5 , 93.0)
					(1 , 92.5)
					(0.5 , 91.05)
					(0.2 , 86.81)
					(0.1 , 81.32)
				};
				
			\end{axis}
		\end{scope}
	\end{tikzpicture}
\end{subfigure}%
\hspace{-8pt}
\begin{subfigure}
	\centering
	\begin{tikzpicture}
		\begin{scope}[scale=0.9]
			\begin{axis}[
				xlabel=Target (\%),
				xmode=log,
				ymin=0,
				ymax=100,
				xtick={90, 10, 1, 0.1},
				xticklabels={10,90,99,99.9},
				mark size = 1pt,
				x dir=reverse,
				ylabel=Accuracy (\%),
				ylabel shift = -6 pt,
				width=0.498\textwidth,
				height=1.75in,
				]
				\addplot coordinates
				{
					(90 , 94.83)
					(80 , 94.88)
					(70 , 94.88)
					(60 , 94.91)
					(50 , 94.92)
					(40 , 94.29)
					(30 , 93.24)
					(25 , 92.08)
					(20 , 84.20)
					(15 , 77.18)
					(10 , 71.01)
					(7.5 , 10)
					(5 , 10)
					(4 , 10)
					(3 , 10)
					(2.5 , 10)
					(2 , 10)
					(1.5 , 10)
					(1 , 10)
					(0.5 , 10)
					(0.2 , 10)
					(0.1 , 10)
				};
				\addplot coordinates
				{
					(90 , 95.10)
					(80 , 95.39)
					(70 , 95.53)
					(60 , 95.32)
					(50 , 94.31)
					(40 , 95.02)
					(30 , 94.98)
					(25 , 94.67)
					(20 , 94.45)
					(15 , 94.36)
					(10 , 93.42)
					(7.5 , 92.94)
					(5 , 91.14)
					(4 , 89.80)
					(3 , 89.25)
					(2.5 , 87.92)
					(2 , 88.00)
					(1.5 , 74.97)
					(1 , 57.99)
					(0.5 , 10)
					(0.2 , 10)
					(0.1 , 10)
				};
				\addplot coordinates
				{
					(80 , 95.38)
					(70 , 95.48)
					(60 , 95.44)
					(50 , 94.96)
					(40 , 94.93)
					(30 , 94.64)
					(25 , 94.25)
					(20 , 94.15)
					(15 , 94.27)
					(10 , 93.72)
					(7.5 , 93.06)
					(5 , 91.93)
					(4 , 91.67)
					(3 , 90.57)
					(2.5 , 89.90)
					(2 , 89.07)
					(1.5 , 87.68)
					(1 , 84.66)
					(0.5 , 79.86)
					(0.2 , 70.18)
					(0.1 , 66.96)
				};
			\end{axis}
		\end{scope}
	\end{tikzpicture}
	\vspace{6pt}
\end{subfigure}
\hspace{100pt} \ref{named}
		\caption{{Comparison} 
 of the trade-off between pruning target and top-1 accuracy for ResNet-20 (with an initial embedding of 64 feature-maps) on CIFAR-10, for SWD and two reference methods. ``Magnitude pruning'' refers either to the method used in Han~et~al.~\cite{han2015learning} or Liu~et~al.~\cite{liu2017learning}. SWD has a better performance/parameter trade-off on high-pruning targets. (\textbf{a}) Unstructured 
 pruning. (\textbf{b})~Structured pruning.}
		\label{tradeoffbigfigurecifar10}
	\end{figure}

\begin{figure}[H]

			\pgfplotsset{footnotesize,samples=10}
			\hspace{-60pt}
\begin{subfigure}{}
	\begin{tikzpicture}
		\begin{scope}[scale=0.85]
			\begin{axis}[
				xlabel=Target (\%),
				xmode=log,
				ymin=0,
				ymax=100,
				mark size = 1pt,
				xtick={90, 10, 1, 0.1},
				xticklabels={10,90,99,99.9},
				x dir=reverse,
				ylabel=Accuracy (\%),
				ylabel shift = -6 pt,
				width=0.498\textwidth,
				height=1.75in,
				legend entries={Magnitude pruning, LR-Rewinding, SWD},
				legend to name=named,
				legend columns=-1,
				legend style = {draw=none},
				]
				
				\addplot coordinates
				{
					(90 , 92.23)
					(80 , 92.25)
					(70 , 92.27)
					(60 , 92.31)
					(50 , 92.43)
					(40 , 91.95)
					(30 , 91.78)
					(25 , 91.46)
					(20 , 90.77)
					(15 , 90.22)
					(10 , 85.26)
					(7.5 , 79.98)
					(5 , 77.15)
					(4 , 79.41)
					(3 , 68.85)
					(2.5 , 68.51)
					(2 , 58.15)
					(1.5 , 41.60)
					(1 , 41.26)
					(0.5 , 34.88)
					(0.2 , 10.0)
					(0.1 , 10.0)
				};
				\addplot coordinates
				{
					(90 , 90.47)
					(80 , 89.7)
					(70 , 92.57)
					(60 , 90.15)
					(50 , 91.06)
					(40 , 89.93)
					(30 , 89.8)
					(25 , 89.39)
					(20 , 91.52)
					(15 , 88.51)
					(10 , 88.12)
					(7.5, 86.07)
					(5 , 83.27)
					(4 , 82.96)
					(3 , 82.75)
					(2.5 , 79.32)
					(2 , 75.21)
					(1.5 , 62.52)
					(1 , 51.93)
					(0.5 , 37.22)
					(0.2 , 10)
					(0.1 , 10)
				};
				\addplot coordinates
				{(90 , 92.63)
					(80 , 92.47)
					(70 , 92.45)
					(60 , 92.36)
					(50 , 92.08)
					(40 , 92.15)
					(30 , 91.69)
					(25 , 90.90)
					(20 , 91.37)
					(15 , 90.97)
					(10 , 90.15)
					(7.5, 89.88)
					(5 , 88.9)
					(4 , 88.69)
					(3 , 86.95)
					(2.5 , 86.16)
					(2 , 84.88)
					(1.5 , 83.33)
					(1 , 75.89)
					(0.5 , 29.35)
					(0.2 , 16.47)
					(0.1 , 11.11)
				};
				
			\end{axis}
		\end{scope}
	\end{tikzpicture}
	\vspace{5pt}
\end{subfigure}%
\hspace{-5pt}
\begin{subfigure}{}
	\begin{tikzpicture}
		\begin{scope}[scale=0.85]
			\begin{axis}[
				xlabel=Target (\%),
				xmode=log,
				ymin=0,
				ymax=100,
				xtick={90, 10, 1, 0.1},
				xticklabels={10,90,99,99.9},
				mark size = 1pt,
				x dir=reverse,
				ylabel=Accuracy (\%),
				ylabel shift = -6 pt,
				width=0.5\textwidth,
				height=1.75in,
				]
				\addplot coordinates
				{
					(90 , 91.96)
					(80 , 91.25)
					(70 , 90.55)
					(60 , 89.59)
					(50 , 89.11)
					(40 , 87.70)
					(30 , 85.08)
					(25 , 82.61)
					(20 , 79.71)
					(15 , 10)
					(10 , 10)
					(7.5 , 10)
					(5 , 64.23)
					(4 , 10)
					(3 , 10)
					(2.5 , 10)
					(2 , 10)
					(1.5 , 10)
					(1 , 10)
					(0.5 , 10)
					(0.2 , 10)
					(0.1 , 10)
				};
				\addplot coordinates
				{
					(90 , 89.95)
					(80 , 88.91)
					(70 , 90.65)
					(60 , 89.94)
					(50 , 88.07)
					(40 , 85.84)
					(30 , 85.84)
					(25 , 83.58)
					(20 , 83.50)
					(15 , 82.53)
					(10 , 80.19)
					(7.5 , 74.81)
					(5 , 66.30)
					(4 , 64.90)
					(3 , 39.30)
					(2.5 , 10)
					(2 , 10)
					(1.5 , 10)
					(1 , 10)
					(0.5 ,10 )
					(0.2 , 10)
					(0.1 , 10)
				};
				\addplot coordinates
				{
					(90 , 91.88)
					(80 , 91.97)
					(70 ,91.22)
					(60 , 90.67)
					(50 , 90.27)
					(40 , 89.66)
					(30 , 88.93)
					(25 , 88.23)
					(20 , 87.82)
					(15 , 86.79)
					(10 , 85.25)
					(7.5 , 83.67)
					(5 , 80.66)
					(4 , 78.39)
					(3 , 75.45)
					(2.5 , 72.73)
					(2 , 71.45)
					(1.5 ,66.71)
					(1 , 51.49)
					(0.5 , 47.63)
					(0.2 , 36.67)
					(0.1 , 10.00)
				};
			\end{axis}
		\end{scope}
	\end{tikzpicture}
		\vspace{6pt}
\end{subfigure}
\hspace{100pt} \ref{named}
		\caption{{Comparison} of the trade-off between pruning target and top-1 accuracy for ResNet-20 on CIFAR-10, with an initial embedding of 16 feature maps, for SWD and two reference methods. ``Magnitude pruning'' refers either to the method used in Han~et~al.~\cite{han2015learning} or Liu~et~al.~\cite{liu2017learning}. SWD has a better performance/parameter trade-off on high-pruning targets. (\textbf{a}) Unstructured pruning. (\textbf{b})~Structured pruning.}
		\label{tradeOff16fm}
	\end{figure}
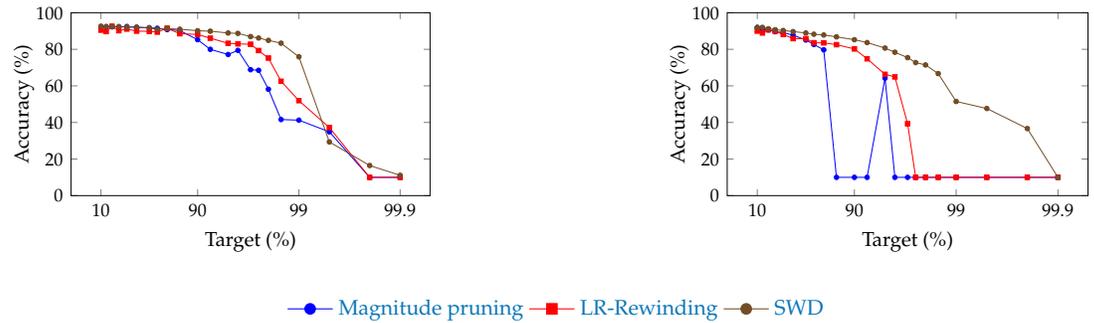

	Figures~\ref{tradeoffbigfigurecifar10} and~\ref{tradeOff16fm} show the influence of SWD on the pruning/accuracy trade-off for ResNet-20, with  an initial embedding of 64 and 16 feature maps, respectively, on CIFAR-10~\cite{krizhevsky2009learning}. We used that lighter dataset instead of ImageNet ILSVRC2012 because of the high cost of computing so many points.
	
		Since each point is the result of only one experiment, there may be some fluctuations due to low statistical power. However, since the same random seed and model initialization were used each time, these may not prevent us from drawing conclusions about the behavior of each method. As we stated in Section~\ref{birth}, pruning originally served as a method to improve generalization. This suggests that the relationship between performance and pruning may be subtle enough to lead to local optima that may not be possible to predict.
	
	\vspace{-6pt}
	
	The standard weight decay (parameter $\mu$) is set to {$5\times 10^{-4}$} for CIFAR-10. The models on CIFAR-10 were trained for 300 epochs and each fine-tuning lasted 15 epochs, except the last one (or the only one when applying Liu~et~al.~\cite{liu2017learning}), which lasted 50 epochs. When using the smooth-$L_1$ norm, its coefficient is set to {$\lambda=1\times 10^{-4}$}. For an initial embedding of 64 feature maps with unstructured pruning, we set {$a_{min}=1\times 10^{-1}$} and {$a_{max}=1\times 10^{5}$}; on structured pruning, {$a_{min}=1\times 10^{2}$} and {$a_{max}=1\times 10^{7}$}. For 16 feature maps, we set $a_{min}=1$ and {$a_{min}=1\times 10^{4}$} for unstructured pruning and $a_{min}=100$ and {$a_{min}=1\times 10^{6}$} when structured. 
	
	Exact results are reported in Table~\ref{unstructured}, in which the expected compression ratios, in terms of operations, are also displayed. Since unstructured pruning produces sparse matrices, whereas structured pruning leads to networks of smaller sizes, some authors such as \mbox{Ma et al.~\cite{ma2021non}} have argued against the use of the former and in favor of the latter. Indeed, sparse matrices either need specific hardware or expensive indexing methods, {which} makes them less efficient than structured pruning. Therefore, because of how hardware- or method-specific the gains of unstructured pruning can be, we preferred not to indicate any compression ratio in terms of operation count for unstructured pruning. However, concerning structured pruning, it is far easier to guess what the operation count will be. The operations are calculated in the following way, with $f_{in}$ being the number of input channels, $f_{out}$ being the number of output channels, $k$ being the kernel size, $h$ being the height (in pixels) of the input feature maps, and $w$ being its width:
		\begin{itemize}
			\item convolution layer: $f_{in}\times f_{out}\times k^2\times h\times w$;
			\item batch normalization layer: $f_{in}\times h\times w \times 2$;
			\item dense layer: $f_{in}\times f_{out} + f_{out}$.
		\end{itemize}
		
		We make no distinction between multiplications and additions in our count.
		
			\subsection{Grid Search on Multiple Models and Datasets}\label{sectionGridsearch}

	To show the influence of the values of $a_{min}$ and $a_{max}$ on the performance of networks right before and right after the final pruning step, we conducted a grid search using LeNet-5 and ResNet-20 on MNIST and CIFAR-10 with both unstructured and structured pruning.
	
	The LeNet-5 models were trained for 200 epochs, with a learning rate of 0.1 and no weight decay (even though $\mu$ is set to {$5\times 10^{-4}$} for SWD); the momentum is set to 0. The pruning targets are 90\% and 99\%. The results of these grid searches are reported in \mbox{Table~\ref{MnistGridSearch}}.
	
	Another grid search, on CIFAR-10 with ResNet-20 (64 channels), is reported in \mbox{Table~\ref{ExtendedGridSearch}} with an extended range of values explored in order to showcase the importance of the increase of $a$ during training. As it involved testing cases decreasing $a$, we named the start and end values of $a$ as $a_{start}$ and $a_{end}$ instead of $a_{min}$ and $a_{max}$. Otherwise, the conditions were the same as described in Sections~\ref{conditions} and \ref{cifarConditions}. Table~\ref{structuredGridSearch} shows another distinct grid search, performed with structured pruning on various pruning targets.
	
		In order to tease apart the sensitivity of SWD from variations of the model or of the dataset, we provide additional grid searches in Tables~\ref{gridsearchUnstructured}~and~\ref{gridsearchStructured}. These tables feature results on CIFAR-10 with ResNet-18 and ResNet-20 to showcase the influence of the model's depth, and on CIFAR-100 with ResNet-34 to have results on another, more complex dataset. Each network has an initial embedding of 64 and we show results for both structured and unstructured pruning.
	
	Additionally, as highlighted by both Tables~\ref{MnistGridSearch}~and~\ref{structuredGridSearch}, the choice of $a_{min}$ and $a_{max}$ depends on the pruning target. To highlight this fact, we show a complete trade-off figure for various values of $a_{min}$ and $a_{max}$ in Figure~\ref{gridTradeoffFig}, whose results are reported in Table~\ref{gridTradeoff}.
	
	\begin{table}[H]

		\caption{Top-1 
 accuracy of ResNet-20, with an initial embedding of 64 or 16 feature maps, on CIFAR-10 for various pruning targets, with different unstructured and structured pruning methods. In both cases, SWD outperforms the other methods for high-pruning targets. For each point, the corresponding estimated percentage of remaining operations (``Ops'') is given (except for unstructured pruning). The missing point in the table (*) is due to the fact that too high values of SWD can lead to overflow of the value of the gradient, which induced a critical failure of the training process on this specific point. However, if the value of $a_{max}$ is instead set to {$1\times 10^{6}$}, we obtain  95.19\% accuracy, with a compression rate of operations of 82.21\%. {The best performance for each target is indicated in bold. Operations are reported in light grey for readability reasons.}}
		\label{unstructured}

\footnotesize
\begin{adjustwidth}{-\extralength}{0cm}
\begin{minipage}{\fulllength}
\begin{tabular}{m{2.5cm}<{\centering} cm{1.1cm}<{\centering}cm{1.1cm}<{\centering}cm{1.1cm}<{\centering} cm{1.1cm}<{\centering}cm{1.1cm}<{\centering}cm{1.1cm}<{\centering}}
	
	\toprule
	\multicolumn{13}{c}{\textbf{ResNet-20 on CIFAR-10 (Unstructured)}}\\
	
	\multicolumn{1}{c}{\textbf{Target}} & 
	\multicolumn{1}{c}{\textbf{Base}} & \multicolumn{1}{c}{\textbf{Ops}} &
	\multicolumn{1}{c}{\textbf{LRR}} & \multicolumn{1}{c}{\textbf{Ops}} &
	\multicolumn{1}{c}{\textbf{SWD}} & \multicolumn{1}{c}{\textbf{Ops}} &
	\multicolumn{1}{c}{\textbf{Base}} & \multicolumn{1}{c}{\textbf{Ops}} &
	\multicolumn{1}{c}{\textbf{LRR}} & \multicolumn{1}{c}{\textbf{Ops}} &
	\multicolumn{1}{c}{\textbf{SWD}} & \multicolumn{1}{c}{\textbf{Ops}}\\
	
	\multicolumn{1}{c}{} & \multicolumn{6}{c}{\textbf{64 Feature Maps}}& \multicolumn{6}{c}{\textbf{16 Feature Maps}}\\
	
	\midrule

	10& \textbf{95.45}&&	94.82	&&95.43&&
	92.23&&
	90.47&&
	\textbf{92.63}&\\
	
	20& 95.47&&	95.15	&&\textbf{95.55}&&
	92.25&&
	89.70&&
	\textbf{92.47}&\\
	
	30& 95.43&&	95.03	&&\textbf{95.47}&&
	92.27&&
	\textbf{92.57}&&
	92.45&\\
	
	40& \textbf{95.48}&&	94.94	&&95.40&&
	92.31&&
	90.15&&
	\textbf{92.36}&\\
	
	50& 95.44&&	95.33	&&\textbf{95.46}&&
	\textbf{92.43}&&
	91.06&&
	92.08&\\
	
	60& 95.32&&	95.04	&&\textbf{95.37}&&
	91.95&&
	89.93&&
	\textbf{92.15}&\\
	
	70& 95.3&&	\textbf{95.45}	&&95.04&&
	\textbf{91.78}&&
	89.8&&
	91.69&\\
	
	75& 95.32&&	95.15	&&\textbf{95.34}&&
	\textbf{91.46}&&
	89.39&&
	90.90&\\
	
	80& \textbf{95.32}&&	95.14	&&95.09&&
	90.77&&
	\textbf{91.52}&&
	91.37&\\
	
	85& \textbf{95.05}&&	95.03	&&94.99&&
	90.22&&
	88.51&&
	\textbf{90.97}&\\
	
	90& 94.77&&	94.72	&&\textbf{94.90}&&
	85.26&&
	88.12&&
	\textbf{90.15}&\\
	
	92.5& 94.48&&	\textbf{94.74}	&&94.58&&
	79.98&&
	86.07&&
	\textbf{89.88}&\\
	
	95& 94.03&&	93.66	&&\textbf{94.40}&&
	77.15&&
	83.27&&
	\textbf{88.90}&\\
	
	96& 93.38&&	93.63	&&\textbf{94.14}&&
	79.41&&
	82.96&&
	\textbf{88.69}&\\
	
	97& 91.95&&	93.34	&&\textbf{93.76}&&
	68.85&&
	82.75&&
	\textbf{86.95}&\\

	97.5& 91.43&&	92.48	&&\textbf{93.52}&&
	68.51&&
	79.32&&
	\textbf{86.16}&\\
	
	98& 90.58&&	91.64	&&\textbf{93.49}&&
	58.15&&
	75.21&&
	\textbf{84.88}&\\
	
	98.5& 87.44&&	90.36	&&\textbf{93.00}&&
	41.60&&
	62.52&&
	\textbf{83.33}&\\
	
	99& 83.42&&	87.38	&&\textbf{92.50}&&
	41.26&&
	51.93&&
	\textbf{75.89}&\\
	
	99.5& 66.90&&	82.21	&&\textbf{91.05}&&
	34.88&&
	\textbf{37.22}&&
	29.35&\\
	
	99.8& 48.52&&	65.46	&&\textbf{86.81}&&
	10.00&&
	10.00&&
	\textbf{16.47}&\\
	
	99.9& 27.78&&	45.44	&&\textbf{81.32}&&
	10.00&&
	10.00&&
	\textbf{11.11}&\\

	\midrule
	
	\multicolumn{13}{c}{\textbf{ResNet-20 on CIFAR-10 (structured)}}\\

	\midrule

	10	&	
	94.83& \textcolor{gray}{84.13}	&
	95.10& \textcolor{gray}{85.37}	&
	*	&\textcolor{gray}{*}&
	\textbf{91.96} & \textcolor{gray}{90.06}&
	89.95 & \textcolor{gray}{85.70}&
	91.88&\textcolor{gray}{77.68}\\
	
	20	& 
	94.88	&  \textcolor{gray}{70.41}& 
	\textbf{95.39} & \textcolor{gray}{76.45}	&
	95.38	& \textcolor{gray}{70.21} &
	91.25 & \textcolor{gray}{78.64}&
	88.91 & \textcolor{gray}{76.21}&
	\textbf{91.97} & \textcolor{gray}{64.63}\\
	
	30	& 
	94.88	&  \textcolor{gray}{58.20}& 
	\textbf{95.53} & \textcolor{gray}{67.45}	&
	95.48	&\textcolor{gray}{59.67} &
	90.55 & \textcolor{gray}{69.77}&
	90.65 & \textcolor{gray}{63.10}&
	\textbf{91.22} & \textcolor{gray}{59.23}\\
	
	40	& 
	94.91	&  \textcolor{gray}{48.06}& 
	95.32 & \textcolor{gray}{53.40}	&
	\textbf{95.44}	&\textcolor{gray}{51.96}&
	89.59 & \textcolor{gray}{62.20}&
	89.94 & \textcolor{gray}{54.85}&
	\textbf{90.67} & \textcolor{gray}{51.07}\\
	
	50	& 
	94.92	&  \textcolor{gray}{40.25} & 
	94.31 & \textcolor{gray}{43.68}	&
	\textbf{94.96}	&\textcolor{gray}{44.31}&
	89.11 & \textcolor{gray}{51.72}&
	88.07 & \textcolor{gray}{44.04}&
	\textbf{90.27} & \textcolor{gray}{41.95}\\
	
	60	& 
	94.29	&  \textcolor{gray}{33.88}& 
	\textbf{95.02} & \textcolor{gray}{35.82}	&
	94.93	& \textcolor{gray}{37.90}&
	87.70 & \textcolor{gray}{42.16}&
	85.84 & \textcolor{gray}{35.6}&
	\textbf{89.66} & \textcolor{gray}{33.42}\\
	
	70	& 
	93.24	&  \textcolor{gray}{26.01} & 
	\textbf{94.98} & \textcolor{gray}{28.08}	&
	94.64	& \textcolor{gray}{30.20}&
	85.08 & \textcolor{gray}{33.12}&
	85.84 & \textcolor{gray}{30.29}&
	\textbf{88.93} & \textcolor{gray}{28.76}\\
	
	75	& 
	92.08	&  \textcolor{gray}{21.36}& 
	\textbf{94.67} & \textcolor{gray}{24.26}	&
	94.25	&\textcolor{gray}{24.89}&
	82.61 & \textcolor{gray}{28.68}&
	83.58 & \textcolor{gray}{22.92}&
	\textbf{88.23} & \textcolor{gray}{25.59}\\
	
	80	& 
	84.20	&  \textcolor{gray}{16.55} & 
	\textbf{94.45} & \textcolor{gray}{19.97}	&
	94.15	&\textcolor{gray}{22.46}&
	79.71 & \textcolor{gray}{24.18}&
	83.50 & \textcolor{gray}{19.07}&
	\textbf{87.82} & \textcolor{gray}{23.94}\\
	
	85	& 
	77.18	&  \textcolor{gray}{12.07} & 
	\textbf{94.36} & \textcolor{gray}{16.45}	&
	94.27	&\textcolor{gray}{19.02}&
	10.00 & \textcolor{gray}{17.29}&
	82.53 & \textcolor{gray}{18.36}&
	\textbf{86.79} & \textcolor{gray}{19.20}\\
	
	90	& 
	71.01	&  \textcolor{gray}{8.04}& 
	93.42 & \textcolor{gray}{11.67}	&
	\textbf{93.72}	& \textcolor{gray}{14.35}&
	10.00 & \textcolor{gray}{12.31}&
	80.19 & \textcolor{gray}{12.06}&
	\textbf{85.25} & \textcolor{gray}{15.75}\\
	
	92.5	& 
	10.00	&   \textcolor{gray}{5.87}& 
	92.94 & \textcolor{gray}{8.93}&
	\textbf{93.06}	&\textcolor{gray}{12.84}&
	10.00 & \textcolor{gray}{10.40}&
	74.81 & \textcolor{gray}{9.86}&
	\textbf{83.67} & \textcolor{gray}{12.38}\\
	
	95	& 
	10.00	&   \textcolor{gray}{4.01}& 
	91.14 & \textcolor{gray}{6.66}&
	\textbf{91.93}	&\textcolor{gray}{9.65}&
	64.23 & \textcolor{gray}{8.1}&
	66.30 & \textcolor{gray}{5.89}&
	\textbf{80.66} & \textcolor{gray}{11.08}\\
	
	96	& 
	10.00	&   \textcolor{gray}{3.39}&
	89.80 & \textcolor{gray}{5.72}&
	\textbf{91.67}	&\textcolor{gray}{8.89}&
	10.00 & \textcolor{gray}{7.16}&
	64.90 & \textcolor{gray}{4.81}&
	\textbf{78.39} & \textcolor{gray}{9.99}\\
	
	97	& 
	10.00	&   \textcolor{gray}{2.84}& 
	89.25 & \textcolor{gray}{4.68}&
	\textbf{90.57}	&\textcolor{gray}{7.45}&
	10.00 & \textcolor{gray}{5.08}&
	39.30 & \textcolor{gray}{4.25}&
	\textbf{75.45} & \textcolor{gray}{8.53}\\
	
	97.5	& 
	10.00	&   \textcolor{gray}{2.26}&
	87.92 & \textcolor{gray}{4.27}&
	\textbf{89.90}	&\textcolor{gray}{7.05}&
	10.00 & \textcolor{gray}{4.21}&
	10.00 & \textcolor{gray}{3.79}&
	\textbf{72.73} & \textcolor{gray}{7.8}\\
	
	98	& 
	10.00	&   \textcolor{gray}{1.80}&
	88.00 & \textcolor{gray}{3.63}&
	\textbf{89.07}	&\textcolor{gray}{6.00}&
	10.00 & \textcolor{gray}{3.7}&
	10.00 & \textcolor{gray}{3.12}&
	\textbf{71.45} & \textcolor{gray}{6.73}\\
	
	98.5	& 
	10.00	&   \textcolor{gray}{1.37}& 
	74.97 & \textcolor{gray}{2.73}&
	\textbf{87.68}	&\textcolor{gray}{5.29}&
	10.00 & \textcolor{gray}{2.13}&
	10.00 & \textcolor{gray}{2.64}&
	\textbf{66.71} & \textcolor{gray}{5.08}\\
	
	99	& 
	10.00	&   \textcolor{gray}{0.99}&
	57.99 & \textcolor{gray}{2.32}&
	\textbf{84.66}	&\textcolor{gray}{4.22}&
	10.00 & \textcolor{gray}{1.79}&
	10.00 & \textcolor{gray}{2.24}&
	\textbf{51.49} & \textcolor{gray}{4.25}\\
	
	99.5	& 
	10.00	&   \textcolor{gray}{0.57}&
	10.00 & \textcolor{gray}{1.45}	&
	\textbf{79.86}	&\textcolor{gray}{2.42}&
	10.00 & \textcolor{gray}{0.8}&
	10.00 & \textcolor{gray}{1.53}&
	\textbf{47.63} & \textcolor{gray}{1.9}\\
	
	99.8	& 
	10.00	&   \textcolor{gray}{0.26}&
	10.00 & \textcolor{gray}{0.75}&
	\textbf{70.18}	&\textcolor{gray}{0.97}&
	10.00 & \textcolor{gray}{0.03}&
	10.00 & \textcolor{gray}{0.13}&
	\textbf{36.67} & \textcolor{gray}{0.53}\\
	
	99.9	& 
	10.00	&   \textcolor{gray}{0.12}&
	10.00 & \textcolor{gray}{0.37}&
	\textbf{66.96}	&\textcolor{gray}{0.45}&
	10.00 & \textcolor{gray}{0.01}&
	10.00 & \textcolor{gray}{0.01}&
	10.00 & \textcolor{gray}{0.35}\\

	\bottomrule
	
\end{tabular}
\end{minipage}
\end{adjustwidth}
	\end{table}

	\begin{table}[H]

		\caption{{Top-1} accuracy after the final unstructured removal step and the difference of performance it induces, for LeNet-5 on MNIST with pruning targets of 10\% and 1\%. We observe that sufficiently high values of $a_{max}$ are needed to prevent the post-removal drop in performance. Higher values of $a_{min}$ seem to work better than smaller ones. The difference induced by $a_{min}$ and $a_{max}$ seems to be more dramatic for higher pruning targets. {Colors are added to ease the interpretation of the results.}}
		\label{MnistGridSearch}
	

\begin{adjustwidth}{-\extralength}{0cm}

\setlength{\tabcolsep}{30mm}
\begin{tabular}{@{} p{0.3cm} c @{\hspace{1mm}} c @{}}
	\toprule
	&\multicolumn{2}{c}{\small\textbf{Grid Search with LeNet-5 on MNIST}}\\
	\midrule
	&\small\textbf{Top-1 Accuracy after Removal (\%)} & \small\textbf{Change of Accuracy through Removal (\%)} \\

	\begin{tikzpicture}[xscale=0.955, yscale=0.5, font=\small]
		\foreach \y [count=\n] in {
			{$a_{min}$}, 
		} {
			\foreach \x [count=\m] in \y {
				\node[minimum width=9.5mm, minimum height = 5mm, text=black] at (\m,-\n) {\x};
			}
		}
	\end{tikzpicture}&\begin{tikzpicture}[xscale=1.5, yscale=0.5, font=\small]
		\foreach \y [count=\n] in {
			{{$1\times 10^{-1}$},    {$1\times 10^{-2}$},    {$1\times 10^{-3}$},    {$1\times 10^{-4}$}}, 
		} {
			\foreach \x [count=\m] in \y {
				\node[minimum width=15mm, minimum height = 5mm, text=black] at (\m,-\n) {\x};
			}
		}
	\end{tikzpicture} & \begin{tikzpicture}[xscale=1.5, yscale=0.5, font=\small]
		\foreach \y [count=\n] in {
			{{$1\times 10^{-1}$},    {$1\times 10^{-2}$},    {$1\times 10^{-3}$},    {$1\times 10^{-4}$}}, 
		} {
			\foreach \x [count=\m] in \y {
				\node[minimum width=15mm, minimum height = 5mm, text=black] at (\m,-\n) {\x};
			}
		}
	\end{tikzpicture} \\
	
	\midrule
	\small$a_{max}$&\multicolumn{2}{c}{\small\textbf{Pruning target 90\%}}\\

	\begin{tikzpicture}[yscale=0.5, font=\small]
		\foreach \y [count=\n] in {
			{{$1\times 10^{1}$}},    {{$1\times 10^{2}$}},   {{$1\times 10^{3}$}},    {{$1\times 10^{4}$}}, 
		} {
			\foreach \x [count=\m] in \y {
				\node[minimum height = 5mm, text=black] at (\m,-\n) {\x};
			}
		}
	\end{tikzpicture}&\begin{tikzpicture}[xscale=1.5, yscale=0.5, font=\small]
		\foreach \y [count=\n] in {
			{98.26,    98.03,    98.02,    95.64}, 
			{98.65,    98.81,    98.74,    98.84},
			{98.81,    98.47,    97.96,    98.55},
			{98.95,    99.04,    96.92,    98.88},
		} {
			\foreach \x [count=\m] in \y {
				\def\p{30}
				\def\min{10.28}
				\def\max{99.06}
				\pgfkeys{/pgf/fpu=true, /pgf/fpu/output format=fixed}
				\pgfmathparse{((\x-\min)/(\max - \min))^\p*100}\edef\j{\pgfmathresult}
				\pgfkeys{/pgf/fpu=false}
				\node[fill=green!\j!red, fill opacity=0.5, text opacity=1, minimum width=15mm, minimum height = 5mm, text=black] at (\m,-\n) {\x};
			}
		}
	\end{tikzpicture} & \begin{tikzpicture}[xscale=1.5, yscale=0.5, font=\small]
		\foreach \y [count=\n] in {
			{-0.45,    -0.58,   -0.73 ,    -3.01}, 
			{0.14,    0.33,    -0.02,    0.11},
			{0.01,    0.12,    -0.02,    0.45},
			{0,    0,    0.37,    -0.03},
		} {
			\foreach \x [count=\m] in \y {
				\def\p{1.5}
				\def\min{-3.56}
				\def\max{0.37}
				\pgfkeys{/pgf/fpu=true, /pgf/fpu/output format=fixed}
				\pgfmathparse{((\x-\min)/(\max - \min))^\p*100}\edef\j{\pgfmathresult}
				\pgfkeys{/pgf/fpu=false}
				\node[fill=green!\j!red, fill opacity=0.5, text opacity=1, minimum width=15mm, minimum height = 5mm, text=black] at ($(\m, -\n)$) {$\x$};
			}
		}
	\end{tikzpicture} \\

	&\multicolumn{2}{c}{\small\textbf{Pruning target 99\%}}\\
	
	\begin{tikzpicture}[yscale=0.5, font=\small]
		\foreach \y [count=\n] in {
			{$1\times 10^{1}$},    {$1\times 10^{2}$},    {$1\times 10^{3}$},    {$1\times 10^{4}$}, 
		} {
			\foreach \x [count=\m] in \y {
				\node[minimum height = 5mm, text=black] at (\m,-\n) {\x};
			}
		}
	\end{tikzpicture}&\begin{tikzpicture}[xscale=1.5, yscale=0.5, font=\small]
		\foreach \y [count=\n] in {
			{23.52,    15.74,    18.04,    12.45}, 
			{19.86,    16.04,    22.94,    22.76},
			{78.06,    75.84,    67.69,    69.74},
			{92.47,    93.52,    92.60,    92.14},
		} {
			\foreach \x [count=\m] in \y {
				\def\p{1}
				\def\min{12.45}
				\def\max{93.52}
				\pgfkeys{/pgf/fpu=true, /pgf/fpu/output format=fixed}
				\pgfmathparse{((\x-\min)/(\max - \min))^\p*100}\edef\j{\pgfmathresult}
				\pgfkeys{/pgf/fpu=false}
				\node[fill=green!\j!red, fill opacity=0.5, text opacity=1, minimum width=15mm, minimum height = 5mm, text=black] at (\m,-\n) {\x};
			}
		}
	\end{tikzpicture} &  \begin{tikzpicture}[xscale=1.5, yscale=0.5, font=\small]
		\foreach \y [count=\n] in {
			{-75.39,    -83.20,    -80.72,    -86.53}, 
			{-70.60,    -80.85,    -74.26,    -72.44},
			{-15.01,    -10.81,    -22.92,    -21.69},
			{2.12,       0.28,       3.03,        2.46},
		} {
			\foreach \x [count=\m] in \y {
				\def\p{1.5}
				\def\min{-86.53}
				\def\max{3.03}
				\pgfkeys{/pgf/fpu=true, /pgf/fpu/output format=fixed}
				\pgfmathparse{((\x-\min)/(\max - \min))^\p*100}\edef\j{\pgfmathresult}
				\pgfkeys{/pgf/fpu=false}
				\node[fill=green!\j!red, fill opacity=0.5, text opacity=1, minimum width=15mm, minimum height = 5mm, text=black] at (\m,-\n) {$\x$};
			}
		}
	\end{tikzpicture} \\
	
	& &\\[-3.2ex]
	
\noalign{\hrule height 1.0pt}
\end{tabular}
\end{adjustwidth}
	\end{table}

	\vspace{-6pt}
	
	\begin{table}[H]
	\footnotesize

		\caption{On ResNet-20 with an initial embedding of 64 feature maps, trained on CIFAR-10 for a pruning target of 90\%. Top-1 accuracy after the final unstructured removal step and the difference in performance it induces. The best results are obtained for reasonably low $a_{start}$ and high $a_{end}$, in accordance with the motivation behind SWD we provided in Section~\ref{swd}. {Colors are added to ease the interpretation of the results.}}
		\label{ExtendedGridSearch}
			
\setlength{\tabcolsep}{20.5mm}
\begin{adjustwidth}{-\extralength}{0cm}
\centering 
\begin{tabular}{@{}c c@{}}
	\toprule
	&\small\textbf{Extended Grid Search}\\
	\begin{tikzpicture}[xscale=1.3, yscale=0.5, font=\small]
		\foreach \y [count=\n] in {
			{$a_{start}$}, 
		} {
			\foreach \x [count=\m] in \y {
				\node[minimum width=13mm, minimum height = 5mm, text=black] at (\m,-\n) {\x};
			}
		}
	\end{tikzpicture}&\begin{tikzpicture}[xscale=1.3, yscale=0.5, font=\small]
		\foreach \y [count=\n] in {
			{{$1\times 10^{4}$}, {$1\times 10^{3}$}, {$1\times 10^{2}$}, {$1\times 10^{1}$}, {$1\times 10^{0}$}, {$1\times 10^{-1}$},    {$1\times 10^{-2}$},    {$1\times 10^{-3}$},    {$1\times 10^{-4}$},    {$1\times 10^{-5}$}}, 
		} {
			\foreach \x [count=\m] in \y {
				\node[minimum width=13mm, minimum height = 5mm, text=black] at (\m,-\n) {\x};
			}
		}
	\end{tikzpicture}\\
	
	\midrule
	\small{$a_{end}$}  & \small{\textbf{Top-1 accuracy after removal (\%)}}\\
	
	\begin{tikzpicture}[xscale=1.3, yscale=0.5, font=\small]
		\foreach \y [count=\n] in {
			{{$1\times 10^{1}$}},    {$1\times 10^{2}$},    {$1\times 10^{3}$},    {$1\times 10^{4}$},    {$1\times 10^{5}$}, 
		} {
			\foreach \x [count=\m] in \y {
				\node[minimum width=13mm, minimum height = 5mm, text=black] at (\m,-\n) {\x};
			}
		}
	\end{tikzpicture}&\begin{tikzpicture}[xscale=1.3, yscale=0.5, font=\small]
		\foreach \y [count=\n] in {
			{94.44,94.08,93.92,94.21,94.54, 91.00,    83.89,    53.68,    71.88,    67.07}, 
			{94.52,94.13,93.91,94.00,94.65,95.15,    94.55,    93.99,    92.40,    89.87},
			{94.57,94.23,93.50,94.00,94.72,94.96,    95.29,    95.27,    94.81,    94.72},
			{94.61,94.17,93.85,94.49,94.50,94.73,    95.37,    95.29,    95.14,    95.22},
			{94.37,94.45,93.54,94.39,94.39,94.78,    95.24,    95.07,    95.30,    95.19},
		} {
			\foreach \x [count=\m] in \y {
				\def\p{20}
				\def\min{67.07}
				\def\max{95.37}
				\pgfkeys{/pgf/fpu=true, /pgf/fpu/output format=fixed}
				\pgfmathparse{((\x-\min)/(\max - \min))^\p*100}\edef\j{\pgfmathresult}
				\pgfkeys{/pgf/fpu=false}
				\node[fill=green!\j!red, minimum width=13mm, fill opacity=0.5, text opacity=1, minimum height = 5mm, text=black] at (\m,-\n) {\x};
			}
		}
	\end{tikzpicture} \\
	
	&\small{\textbf{Change in accuracy through removal (\%)}}\\
	
	\begin{tikzpicture}[xscale=1.3, yscale=0.5, font=\small]
		\foreach \y [count=\n] in {
			{$1\times 10^{1}$},    {$1\times 10^{2}$},    {$1\times 10^{3}$},    {$1\times 10^{4}$},    {$1\times 10^{5}$}, 
		} {
			\foreach \x [count=\m] in \y {
				\node[minimum width=13mm, minimum height = 5mm, text=black] at (\m,-\n) {\x};
			}
		}
	\end{tikzpicture}&\begin{tikzpicture}[xscale=1.3, yscale=0.5, font=\small]
		\foreach \y [count=\n] in {
			{-0.1,0,-0.01,0, 0.06,-4.33,    -11.49,    -41.72,    -23.53,    -28.36}, 
			{0.06,-0.09,-0.01,-0.05,0.1,-0.01,    -0.45,    -0.99,    -2.37,    -4.74},
			{-0.01,0.03,-0.01,0,0.02,-0.01,    -0.06,    0.05,    -0.01,    0.11},
			{-0.02,-0.07,0.01,0.01,0.02,-0.03,    0.02,    0.05,    -0.02,    0.08},
			{0.01,0.04,0.02,0,0,-0.01,    -0.02,    -0.01,    0.01,    -0.10},
		} {
			\foreach \x [count=\m] in \y {
				\def\p{60}
				\def\min{-41.72}
				\def\max{0.06}
				\pgfkeys{/pgf/fpu=true, /pgf/fpu/output format=fixed}
				\pgfmathparse{((\x-\min)/(\max - \min))^\p*100}\edef\j{\pgfmathresult}
				\pgfkeys{/pgf/fpu=false}
				\node[fill=green!\j!red, minimum width=13mm, fill opacity=0.5, text opacity=1, minimum height = 5mm, text=black] at (\m,-\n) {$\x$};
			}
		}
	\end{tikzpicture}\\
		& \\[-3.75ex]
	\bottomrule
\end{tabular}
\end{adjustwidth}
	\end{table}
	\unskip
	\begin{table}[H]
		\footnotesize

		\caption{Top-1 accuracy after the final structured removal step and the difference in performance it induces, for ResNet-20 with an initial embedding of 64 feature maps, trained on CIFAR-10 and pruning targets of 75\% and 90\%. Structured pruning with SWD turned out to require exploring a wider range of values than unstructured pruning, as well as being even more sensitive to $a$. {Colors are added to ease the interpretation of the results.}}
		\label{structuredGridSearch}
\setlength{\tabcolsep}{21mm}
\begin{adjustwidth}{-\extralength}{0cm}
\centering 
\begin{tabular}{@{} p{1cm} c @{\hspace{1mm}} c @{}}
	\toprule
	&\multicolumn{2}{c}{\small\textbf{Grid Search with Structured Pruning}}\\
	&\small\textbf{Top-1 Accuracy after Removal (\%)} & \small\textbf{Change of Accuracy through Removal (\%)} \\

	\begin{tikzpicture}[xscale=1.3, yscale=0.5, font=\small]
		\foreach \y [count=\n] in {
			{$a_{min}$}, 
		} {
			\foreach \x [count=\m] in \y {
				\node[minimum width=13mm, minimum height = 5mm, text=black] at (\m,-\n) {\x};
			}
		}
	\end{tikzpicture}&\begin{tikzpicture}[xscale=1.3, yscale=0.5, font=\small]
		\foreach \y [count=\n] in {	
			{{$1\times 10^{0}$}, {$1\times 10^{-1}$},    {$1\times 10^{-2}$},    {$1\times 10^{-3}$},    {$1\times 10^{-4}$}},
		} {
			\foreach \x [count=\m] in \y {
				\node[minimum width=13mm, minimum height = 5mm, text=black] at (\m,-\n) {\x};
			}
		}
	\end{tikzpicture} & \begin{tikzpicture}[xscale=1.3, yscale=0.5, font=\small]
		\foreach \y [count=\n] in {
			{{$1\times 10^{0}$}, {$1\times 10^{-1}$},    {$1\times 10^{-2}$},    {$1\times 10^{-3}$},    {$1\times 10^{-4}$}},
		} {
			\foreach \x [count=\m] in \y {
				\node[minimum width=13mm, minimum height = 5mm, text=black] at (\m,-\n) {\x};
			}
		}
	\end{tikzpicture} \\

	\midrule
	\small$a_{max}$&\multicolumn{2}{c}{\small\textbf{Pruning target 75\%}}\\

	\begin{tikzpicture}[xscale=1.3, yscale=0.5, font=\small]
		\foreach \y [count=\n] in {
			{$1\times 10^{1}$},    {$1\times 10^{2}$},    {$1\times 10^{3}$},    {$1\times 10^{4}$},    {$1\times 10^{5}$},     {$1\times 10^{6}$}
		} {
			\foreach \x [count=\m] in \y {
				\node[minimum width=13mm, minimum height = 5mm, text=black] at (\m,-\n) {\x};
			}
		}
	\end{tikzpicture}&\begin{tikzpicture}[xscale=1.3, yscale=0.5, font=\small]
		\foreach \y [count=\n] in {
			{57.11, 8.2,	12.68,	9.52,	7.9},
			{91.21, 5.57,	6.45,	9.97,	8.01},
			{94.94, 86.11,	15.87,	5.84,	4.38},
			{94.79, 94.77,	94.24,	36.51,	6.63},
			{94.70, 95.28,	94.75,	94.59,	86.47},
			{95.05, 94.91, 94.70, 94.99, 94.71},
		} {
			\foreach \x [count=\m] in \y {
				\def\p{2}
				\def\min{4.38}
				\def\max{95.28}
				\pgfkeys{/pgf/fpu=true, /pgf/fpu/output format=fixed}
				\pgfmathparse{((\x-\min)/(\max - \min))^\p*100}\edef\j{\pgfmathresult}
				\pgfkeys{/pgf/fpu=false}
				\node[fill=green!\j!red, fill opacity=0.5, text opacity=1, minimum width=13mm, minimum height = 5mm, text=black] at (\m,-\n) {\x};
			}
		}
	\end{tikzpicture} & \begin{tikzpicture}[xscale=1.3, yscale=0.5, font=\small]
		\foreach \y [count=\n] in {
			{-37.68, -87.4, -82.68, -85.81, -87.61}, 
			{-3.74, -89.35, -88.37, -85.06, -86.72},
			{-0.05, -7.78, -77.34, -87.85, -89.39},
			{0, -0.04, -0.21, -56.77, -85.94},
			{-0.01, 0, 0.2, 0.05, -7.83},
			{0, 0, -0.01, 0, 0.01},
		}{
			\foreach \x [count=\m] in \y {
				\def\p{2}
				\def\min{-89.39}
				\def\max{0.05}
				\pgfkeys{/pgf/fpu=true, /pgf/fpu/output format=fixed}
				\pgfmathparse{((\x-\min)/(\max - \min))^\p*100}\edef\j{\pgfmathresult}
				\pgfkeys{/pgf/fpu=false}
				\node[fill=green!\j!red, fill opacity=0.5, text opacity=1, minimum width=13mm, minimum height = 5mm, text=black] at (\m,-\n) {$\x$};
			}
		}
	\end{tikzpicture} \\

	&\multicolumn{2}{c}{\small\textbf{Pruning target 90\%}}\\

	\begin{tikzpicture}[xscale=1.3, yscale=0.5, font=\small]
		\foreach \y [count=\n] in {
			{$1\times 10^{1}$},    {$1\times 10^{2}$},    {$1\times 10^{3}$},    {$1\times 10^{4}$},     {$1\times 10^{5}$},    {$1\times 10^{6}$},
		} {
			\foreach \x [count=\m] in \y {
				\node[minimum width=13mm, minimum height = 5mm, text=black] at (\m,-\n) {\x};
			}
		}
	\end{tikzpicture}&\begin{tikzpicture}[xscale=1.3, yscale=0.5, font=\small]
		\foreach \y [count=\n] in {
			{11.17, 10.00, 10.00, 10.00, 07.19}, 
			{16.45, 10.00, 10.00, 10.00, 10.00},
			{78.66, 14.51, 10.01, 10.00, 10.04},
			{94.15, 92.86, 49.28, 51.28, 10.55},
			{93.82, 93.59, 93.18, 89.49, 91.93},
			{94.05, 94.02, 93.73, 93.60, 93.55},
		} {
			\foreach \x [count=\m] in \y {
				\def\p{1}
				\def\min{10.00}
				\def\max{94.15}
				\pgfkeys{/pgf/fpu=true, /pgf/fpu/output format=fixed}
				\pgfmathparse{((\x-\min)/(\max - \min))^\p*100}\edef\j{\pgfmathresult}
				\pgfkeys{/pgf/fpu=false}
				\node[fill=green!\j!red, fill opacity=0.5, text opacity=1, minimum width=13mm, minimum height = 5mm, text=black] at (\m,-\n) {\x};
			}
		}
	\end{tikzpicture} & \begin{tikzpicture}[xscale=1.3, yscale=0.5, font=\small]
		\foreach \y [count=\n] in {
			{-83.86, -85.19, -85.24, -85.45, -88.22}, 
			{-78.2, -84.05, -84.62, -84.22, -84.11},
			{-14.91, -77.96, -82.53, -81.64, -81.37},
			{-0.01, -0.79, -42.71, -40.7, -81.47},
			{-0.04, -0.01, -0.03, -2.4, -0.39},
			{0, 0, 0.01, 0, -0.02},
		} {
			\foreach \x [count=\m] in \y {
				\def\p{10}
				\def\min{-88.22}
				\def\max{0.01}
				\pgfkeys{/pgf/fpu=true, /pgf/fpu/output format=fixed}
				\pgfmathparse{((\x-\min)/(\max - \min))^\p*100}\edef\j{\pgfmathresult}
				\pgfkeys{/pgf/fpu=false}
				\node[fill=green!\j!red, fill opacity=0.5, text opacity=1, minimum width=13mm, minimum height = 5mm, text=black] at (\m,-\n) {$\x$};
			}
		}
	\end{tikzpicture} \\
	
		& &\\[-3.75ex]
	
	\bottomrule
\end{tabular}
\end{adjustwidth}
	\end{table}
		\vspace{-6pt}
		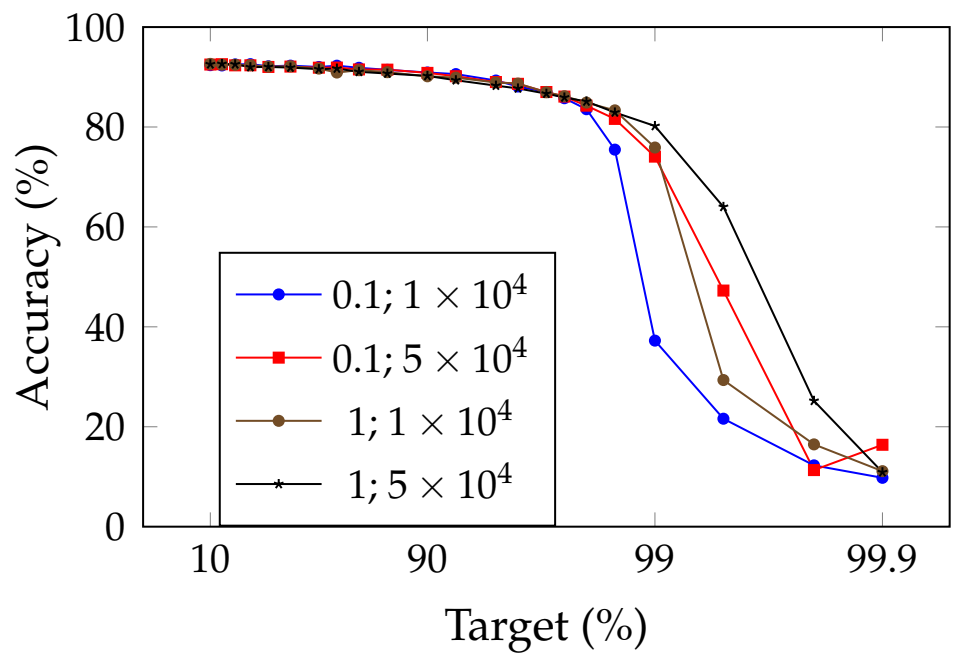
\begin{figure}[H]
				\pgfplotsset{footnotesize,samples=10}
\begin{tikzpicture}
	\begin{scope}[scale=1.9]
		\begin{axis}[
			xlabel=Target (\%),
			xmode=log,
			ymin=0,
			ymax=100,
			mark size = 1pt,
			xtick={90, 10, 1, 0.1},
			xticklabels={10,90,99,99.9},
			x dir=reverse,
			ylabel=Accuracy (\%),
			ylabel shift = -6 pt,
			width=0.50\textwidth,
			height=2in,
			legend entries={0.1; {$1\times 10^{4}$}, 0.1; {$5\times 10^{4}$}, 1; {$1\times 10^{4}$}, 1; {$5\times 10^{4}$}},
			legend style = {at={(axis cs:0,0)},anchor=south east},
			]
			
			\addplot coordinates
			{
				(90.0, 92.38)
				(80.0, 92.32)
				(70.0, 92.53)
				(60.0, 92.58)
				(50.0, 92.15)
				(40.0, 92.28)
				(30.0, 92.01)
				(25.0, 92.27)
				(20.0, 91.85)
				(15.0, 91.44)
				(10.0, 90.91)
				(7.5, 90.59)
				(5.0, 89.3)
				(4.0, 88.11)
				(3.0, 87.01)
				(2.5, 85.76)
				(2.0, 83.56)
				(1.5, 75.47)
				(1.0, 37.24)
				(0.5, 21.61)
				(0.2, 12.27)
				(0.1, 9.78)
			};
			\addplot coordinates
			{
				(90.0, 92.5)
				(80.0, 92.57)
				(70.0, 92.34)
				(60.0, 92.35)
				(50.0, 92.02)
				(40.0, 92.09)
				(30.0, 91.87)
				(25.0, 91.89)
				(20.0, 91.52)
				(15.0, 91.48)
				(10.0, 90.83)
				(7.5, 90.16)
				(5.0, 89.0)
				(4.0, 88.64)
				(3.0, 87.0)
				(2.5, 86.09)
				(2.0, 84.27)
				(1.5, 81.62)
				(1.0, 74.07)
				(0.5, 47.26)
				(0.2, 11.33)
				(0.1, 16.39)
				
			};
			\addplot coordinates
			{(90 , 92.63)
				(80 , 92.47)
				(70 , 92.45)
				(60 , 92.36)
				(50 , 92.08)
				(40 , 92.15)
				(30 , 91.69)
				(25 , 90.90)
				(20 , 91.37)
				(15 , 90.97)
				(10 , 90.15)
				(7.5, 89.88)
				(5 , 88.9)
				(4 , 88.69)
				(3 , 86.95)
				(2.5 , 86.16)
				(2 , 84.88)
				(1.5 , 83.33)
				(1 , 75.89)
				(0.5 , 29.35)
				(0.2 , 16.47)
				(0.1 , 11.11)
			};
			
			\addplot coordinates
			{(90.0, 92.56)
				(80.0, 92.62)
				(70.0, 92.55)
				(60.0, 91.98)
				(50.0, 92.02)
				(40.0, 91.89)
				(30.0, 91.57)
				(25.0, 91.7)
				(20.0, 91.04)
				(15.0, 90.7)
				(10.0, 90.22)
				(7.5, 89.36)
				(5.0, 88.28)
				(4.0, 87.72)
				(3.0, 86.67)
				(2.5, 85.91)
				(2.0, 85.11)
				(1.5, 82.91)
				(1.0, 80.2)
				(0.5, 64.01)
				(0.2, 25.19)
				(0.1, 10.9)
				
			};
			
		\end{axis}
	\end{scope}
\end{tikzpicture}
		\caption{{Comparison} of the trade-off between an unstructured pruning target and the top-1 accuracy for ResNet-20 on CIFAR-10, with an initial embedding of 16 feature maps, for SWD with different values of $a_{min}$ and $a_{max}$. Depending on the pruning rate, the best values to choose are not always the~same.}
		\label{gridTradeoffFig}
	\end{figure}
	
		\begin{table}[H]
	\footnotesize

		\caption{Top-1 accuracy after the final unstructured removal step and the difference in performance it induces, for various networks and datasets with a pruning target of 90\%. The influence of $a_{min}$ and $a_{max}$ varies significantly depending on the problem, although common tendencies persist. {Colors are added to ease the interpretation of the results.}}
		\label{gridsearchUnstructured}
\setlength{\tabcolsep}{20.5mm}

\begin{adjustwidth}{-\extralength}{0cm}
\begin{tabular}{@{} p{1cm} c @{\hspace{1mm}} c @{}}
	\toprule
	&\multicolumn{2}{c}{\small\textbf{Grid Search with Unstructured Pruning}}\\
	&\small\textbf{Top-1 Accuracy after Removal (\%)} & \small\textbf{Change of Accuracy through Removal (\%)} \\

	\begin{tikzpicture}[xscale=1.3, yscale=0.5, font=\small]
		\foreach \y [count=\n] in {
			{$a_{min}$}, 
		} {
			\foreach \x [count=\m] in \y {
				\node[minimum width=13mm, minimum height = 5mm, text=black] at (\m,-\n) {\x};
			}
		}
	\end{tikzpicture}&\begin{tikzpicture}[xscale=1.3, yscale=0.5, font=\small]
		\foreach \y [count=\n] in {
			{{$1\times 10^{-1}$},    {$1\times 10^{-2}$},    {$1\times 10^{-3}$},    {$1\times 10^{-4}$},    {$1\times 10^{-5}$}}, 
		} {
			\foreach \x [count=\m] in \y {
				\node[minimum width=13mm, minimum height = 5mm, text=black] at (\m,-\n) {\x};
			}
		}
	\end{tikzpicture} & \begin{tikzpicture}[xscale=1.3, yscale=0.5, font=\small]
		\foreach \y [count=\n] in {
			{{$1\times 10^{-1}$},    {$1\times 10^{-2}$},    {$1\times 10^{-3}$},    {$1\times 10^{-4}$},    {$1\times 10^{-5}$}}, 
		} {
			\foreach \x [count=\m] in \y {
				\node[minimum width=13mm, minimum height = 5mm, text=black] at (\m,-\n) {\x};
			}
		}
	\end{tikzpicture} \\
	
	\midrule
	\small$a_{max}$&\multicolumn{2}{c}{\small\textbf{ResNet-18 on CIFAR-10}}\\

	\begin{tikzpicture}[yscale=0.5, font=\small]
		\foreach \y [count=\n] in {
			{$1\times 10^{1}$},    {$1\times 10^{2}$},    {$1\times 10^{3}$},    {$1\times 10^{4}$}, {$1\times 10^{5}$},
		} {
			\foreach \x [count=\m] in \y {
				\node[minimum height = 5mm, text=black] at (\m,-\n) {\x};
			}
		}
	\end{tikzpicture}&\begin{tikzpicture}[xscale=1.3, yscale=0.5, font=\small]
		\foreach \y [count=\n] in {
			{94.73, 92.69, 78.28, 83.08, 63.66 },
			{94.88, 94.33, 92.29, 85.33, 29.18 },
			{95.33, 95.34, 93.55, 92.64, 86.89 },
			{95.17, 95.08, 95.28, 94.88, 95.23 },
			{95.19, 95.18, 95.21, 95.11, 95.43 },
		} {
			\foreach \x [count=\m] in \y {
				\def\p{30}
				\def\min{29.18}
				\def\max{95.43}
				\pgfkeys{/pgf/fpu=true, /pgf/fpu/output format=fixed}
				\pgfmathparse{((\x-\min)/(\max - \min))^\p*100}\edef\j{\pgfmathresult}
				\pgfkeys{/pgf/fpu=false}
				\node[fill=green!\j!red, fill opacity=0.5, text opacity=1, minimum width=13mm, minimum height = 5mm, text=black] at (\m,-\n) {\x};
			}
		}
	\end{tikzpicture} & \begin{tikzpicture}[xscale=1.3, yscale=0.5, font=\small]
		\foreach \y [count=\n] in {
			{-0.21, -2.55, -17.3, -12.25, -31.39 },
			{-0.08, -0.68, -2.72, -9.52, -65.73 },
			{0.04, 0.02, -1.64, -2.48, -8.13 },
			{-0.01, 0.02, -0.02, -0.29, 0.01 },
			{0.0, 0.0, 0.01, 0.02, 0.03 },
		} {
			\foreach \x [count=\m] in \y {
				\def\p{5}
				\def\min{-65.73}
				\def\max{0.04}
				\pgfkeys{/pgf/fpu=true, /pgf/fpu/output format=fixed}
				\pgfmathparse{((\x-\min)/(\max - \min))^\p*100}\edef\j{\pgfmathresult}
				\pgfkeys{/pgf/fpu=false}
				\node[fill=green!\j!red, fill opacity=0.5, text opacity=1, minimum width=13mm, minimum height = 5mm, text=black] at (\m,-\n) {$\x$};
			}
		}
	\end{tikzpicture} \\

	&\multicolumn{2}{c}{\small\textbf{ResNet-20 on CIFAR-10}}\\

	\begin{tikzpicture}[yscale=0.5, font=\small]
		\foreach \y [count=\n] in {
			{$1\times 10^{1}$},    {$1\times 10^{2}$},    {$1\times 10^{3}$},    {$1\times 10^{4}$},  {$1\times 10^{5}$},
		} {
			\foreach \x [count=\m] in \y {
				\node[minimum height = 5mm, text=black] at (\m,-\n) {\x};
			}
		}
	\end{tikzpicture}&\begin{tikzpicture}[xscale=1.3, yscale=0.5, font=\small]
		\foreach \y [count=\n] in {
			{91.00, 83.89, 53.68, 71.88, 67.07 },
			{95.15, 94.55, 93.99, 92.40, 89.87 },
			{94.96, 95.29, 95.27, 94.81, 94.72 },
			{94.73, 95.37, 95.29, 95.14, 95.22 },
			{94.78, 95.24, 95.07, 95.30, 95.19 },
		} {
			\foreach \x [count=\m] in \y {
				\def\p{30}
				\def\min{53.68}
				\def\max{95.37}
				\pgfkeys{/pgf/fpu=true, /pgf/fpu/output format=fixed}
				\pgfmathparse{((\x-\min)/(\max - \min))^\p*100}\edef\j{\pgfmathresult}
				\pgfkeys{/pgf/fpu=false}
				\node[fill=green!\j!red, fill opacity=0.5, text opacity=1, minimum width=13mm, minimum height = 5mm, text=black] at (\m,-\n) {\x};
			}
		}
	\end{tikzpicture} & \begin{tikzpicture}[xscale=1.3, yscale=0.5, font=\small]
		\foreach \y [count=\n] in {
			{-4.33, -11.49, -41.72, -23.53, -28.36 },
			{-0.01, -0.45, -0.99, -2.37, -4.74 },
			{-0.01, -0.06, 0.05, -0.01, 0.11 },
			{-0.03, 0.02, 0.05, -0.02, 0.08 },
			{-0.01, -0.02, -0.01, 0.01, -0.1 },
		} {
			\foreach \x [count=\m] in \y {
				\def\p{5}
				\def\min{-41.72}
				\def\max{0.11}
				\pgfkeys{/pgf/fpu=true, /pgf/fpu/output format=fixed}
				\pgfmathparse{((\x-\min)/(\max - \min))^\p*100}\edef\j{\pgfmathresult}
				\pgfkeys{/pgf/fpu=false}
				\node[fill=green!\j!red, fill opacity=0.5, text opacity=1, minimum width=13mm, minimum height = 5mm, text=black] at (\m,-\n) {$\x$};
			}
		}
	\end{tikzpicture} \\

	&\multicolumn{2}{c}{\small\textbf{ResNet-34 on CIFAR-100}}\\

	\begin{tikzpicture}[yscale=0.5, font=\small]
		\foreach \y [count=\n] in {
			{$1\times 10^{1}$},    {$1\times 10^{2}$},    {$1\times 10^{3}$},    {$1\times 10^{4}$},  {$1\times 10^{5}$},
		} {
			\foreach \x [count=\m] in \y {
				\node[minimum height = 5mm, text=black] at (\m,-\n) {\x};
			}
		}
	\end{tikzpicture}&\begin{tikzpicture}[xscale=1.3, yscale=0.5, font=\small]
		\foreach \y [count=\n] in {
			{73.15, 71.94, 66.81, 67.04, 64.55 },
			{77.36, 77.38, 75.71, 75.57, 74.77 },
			{72.63, 78.04, 78.16, 77.52, 77.45 },
			{76.86, 77.55, 78.21, 78.74, 77.52 },
			{77.08, 77.88, 78.12, 77.26, 77.87 },
		} {
			\foreach \x [count=\m] in \y {
				\def\p{2}
				\def\min{64.55}
				\def\max{78.74}
				\pgfkeys{/pgf/fpu=true, /pgf/fpu/output format=fixed}
				\pgfmathparse{((\x-\min)/(\max - \min))^\p*100}\edef\j{\pgfmathresult}
				\pgfkeys{/pgf/fpu=false}
				\node[fill=green!\j!red, fill opacity=0.5, text opacity=1, minimum width=13mm, minimum height = 5mm, text=black] at (\m,-\n) {\x};
			}
		}
	\end{tikzpicture} & \begin{tikzpicture}[xscale=1.3, yscale=0.5, font=\small]
		\foreach \y [count=\n] in {
			{2.5, -6.0, -11.22, -11.16, -13.77 },
			{0.08, -0.46, -1.42, -1.82, -2.55 },
			{1.66, 0.03, -0.02, 0.02, -0.09 },
			{0.08, 0.04, 0.03, 0.01, 0.02 },
			{0.0, -0.01, 0.04, 0.03, 0.0 },
		} {
			\foreach \x [count=\m] in \y {
				\def\p{2}
				\def\min{-13.77}
				\def\max{2.5}
				\pgfkeys{/pgf/fpu=true, /pgf/fpu/output format=fixed}
				\pgfmathparse{((\x-\min)/(\max - \min))^\p*100}\edef\j{\pgfmathresult}
				\pgfkeys{/pgf/fpu=false}
				\node[fill=green!\j!red, fill opacity=0.5, text opacity=1, minimum width=13mm, minimum height = 5mm, text=black] at (\m,-\n) {$\x$};
			}
		}
	\end{tikzpicture} \vspace{-4pt}\\
	
	\bottomrule
\end{tabular}
\end{adjustwidth}
	\end{table}
	
		\vspace{-6pt}

	\begin{table}[H]
				\small

		\caption{Top-1 accuracy after the final structured removal step and the difference in performance it induces, for various networks and datasets with a pruning target of 90\%. The influence of $a_{min}$ and $a_{max}$ varies significantly depending on the problem, although common tendencies persist. As previously shown in Table~\ref{structuredGridSearch}, structured pruning is a lot more sensitive to variations of $a_{min}$ and $a_{max}$. {Colors are added to ease the interpretation of the results.}}
		\label{gridsearchStructured}

\setlength{\tabcolsep}{9mm}
\begin{adjustwidth}{-\extralength}{0cm}
\begin{tabular}{@{} p{0.75cm} c @{\hspace{1mm}} c @{}}
	\toprule
	&\multicolumn{2}{c}{\small\textbf{Grid Search with Structured Pruning}}\\
	&\small\textbf{Top-1 Accuracy after Removal (\%)} & \small\textbf{Change in Accuracy through Removal (\%)} \\
	
	\midrule
	
	\small$a_{min}$&\begin{tikzpicture}[xscale=1.3, yscale=0.5, font=\small]
		\foreach \y [count=\n] in {
			{{$1\times 10^{0}$},    {$1\times 10^{-1}$},    {$1\times 10^{-2}$},    {$1\times 10^{-3}$},    {$1\times 10^{-4}$},    {$1\times 10^{-5}$}}, 
		} {
			\foreach \x [count=\m] in \y {
				\node[minimum width=13mm, minimum height = 5mm, text=black] at (\m,-\n) {\x};
			}
		}
	\end{tikzpicture} & \begin{tikzpicture}[xscale=1.3, yscale=0.5, font=\small]
		\foreach \y [count=\n] in {
			{{$1\times 10^{0}$},    {$1\times 10^{-1}$},    {$1\times 10^{-2}$},    {$1\times 10^{-3}$},    {$1\times 10^{-4}$},    {$1\times 10^{-5}$}}, 
		} {
			\foreach \x [count=\m] in \y {
				\node[minimum width=13mm, minimum height = 5mm, text=black] at (\m,-\n) {\x};
			}
		}
	\end{tikzpicture} \\
	
	\midrule
	\small$a_{max}$&\multicolumn{2}{c}{\small\textbf{ResNet-18 on CIFAR-10}}\\
	
	\begin{tikzpicture}[yscale=0.5, font=\small]
		\foreach \y [count=\n] in {
			{$1\times 10^{1}$},    {$1\times 10^{2}$},    {$1\times 10^{3}$},    {$1\times 10^{4}$}, {$1\times 10^{5}$}, {$1\times 10^{6}$},
		} {
			\foreach \x [count=\m] in \y {
				\node[minimum height = 5mm, text=black] at (\m,-\n) {\x};
			}
		}
	\end{tikzpicture}&\begin{tikzpicture}[xscale=1.3, yscale=0.5, font=\small]
		\foreach \y [count=\n] in {
			{28.95, 10.00, 10.00, 10.00, 10.00, 10.00 },
			{65.74, 10.00, 10.00, 10.00, 10.00, 10.00 },
			{93.81, 10.00, 10.00, 10.00, 10.00, 10.00 },
			{94.63, 18.08, 10.00, 10.00, 10.00, 10.00 },
			{94.84, 94.23, 29.14, 11.39, 10.00, 10.00 },
			{94.73, 94.94, 94.58, 94.46, 93.57, 91.37 },
		} {
			\foreach \x [count=\m] in \y {
				\def\p{30}
				\def\min{10.00}
				\def\max{94.84}
				\pgfkeys{/pgf/fpu=true, /pgf/fpu/output format=fixed}
				\pgfmathparse{((\x-\min)/(\max - \min))^\p*100}\edef\j{\pgfmathresult}
				\pgfkeys{/pgf/fpu=false}
				\node[fill=green!\j!red, fill opacity=0.5, text opacity=1, minimum width=13mm, minimum height = 5mm, text=black] at (\m,-\n) {\x};
			}
		}
	\end{tikzpicture} & \begin{tikzpicture}[xscale=1.3, yscale=0.5, font=\small]
		\foreach \y [count=\n] in {
			{-66.2, -84.93, -84.96, -85.19, -85.19, -84.91 },
			{-28.58, -84.33, -84.52, -84.37, -84.57, -84.33 },
			{-1.05, -84.49, -84.26, -81.58, -84.22, -84.16 },
			{-0.02, -76.53, -84.43, -84.22, -83.7, -79.16 },
			{0.0, -0.62, -65.14, -83.13, -83.69, -82.3 },
			{0.0, 0.01, -0.13, -0.05, 1.33, -2.78 },
		} {
			\foreach \x [count=\m] in \y {
				\def\p{1.5}
				\def\min{-85.19}
				\def\max{1.33}
				\pgfkeys{/pgf/fpu=true, /pgf/fpu/output format=fixed}
				\pgfmathparse{((\x-\min)/(\max - \min))^\p*100}\edef\j{\pgfmathresult}
				\pgfkeys{/pgf/fpu=false}
				\node[fill=green!\j!red, fill opacity=0.5, text opacity=1, minimum width=13mm, minimum height = 5mm, text=black] at (\m,-\n) {$\x$};
			}
		}
	\end{tikzpicture} \\
			& &\\[-3.75ex]
	\bottomrule
\end{tabular}

\end{adjustwidth}
	\end{table}

	\begin{table}[H]\ContinuedFloat
				\small

		\caption{{\em Cont.}}
		\label{gridsearchStructured}

\setlength{\tabcolsep}{9mm}
\begin{adjustwidth}{-\extralength}{0cm}
\begin{tabular}{@{} p{0.75cm} c @{\hspace{1mm}} c @{}}
	\toprule
	&\multicolumn{2}{c}{\small\textbf{Grid Search with Structured Pruning}}\\
	&\small\textbf{Top-1 Accuracy after Removal (\%)} & \small\textbf{Change in Accuracy through Removal (\%)} \\
	
	\midrule
	&\multicolumn{2}{c}{\small\textbf{ResNet-20 on CIFAR-10}}\\
	
	\begin{tikzpicture}[yscale=0.5, font=\small]
		\foreach \y [count=\n] in {
			{$1\times 10^{1}$},    {$1\times 10^{2}$},    {$1\times 10^{3}$},    {$1\times 10^{4}$},  {$1\times 10^{5}$}, {$1\times 10^{6}$},
		} {
			\foreach \x [count=\m] in \y {
				\node[minimum height = 5mm, text=black] at (\m,-\n) {\x};
			}
		}
	\end{tikzpicture}&\begin{tikzpicture}[xscale=1.3, yscale=0.5, font=\small]
		\foreach \y [count=\n] in {
			{11.17, 10.00, 10.00, 10.00, 07.16, 10.17},
			{16.45, 10.00, 10.00, 10.00, 10.00, 10.36 },
			{78.66, 14.51, 10.01, 10.00, 10.04, 11.69 },
			{94.15, 92.86, 49.28, 51.28, 10.55, 15.53 },
			{93.82, 93.59, 93.18, 89.49, 91.93, 89.56 },
			{94.05, 94.02, 93.73, 93.60, 93.55, 92.00 },
		} {
			\foreach \x [count=\m] in \y {
				\def\p{1}
				\def\min{10}
				\def\max{94.05}
				\pgfkeys{/pgf/fpu=true, /pgf/fpu/output format=fixed}
				\pgfmathparse{((\x-\min)/(\max - \min))^\p*100}\edef\j{\pgfmathresult}
				\pgfkeys{/pgf/fpu=false}
				\node[fill=green!\j!red, fill opacity=0.5, text opacity=1, minimum width=13mm, minimum height = 5mm, text=black] at (\m,-\n) {\x};
			}
		}
	\end{tikzpicture} & \begin{tikzpicture}[xscale=1.3, yscale=0.5, font=\small]
		\foreach \y [count=\n] in {
			{-83.86, -85.19, -85.24, -85.45, -88.22, -85.31 },
			{-78.2, -84.05, -84.62, -84.22, -84.11, -83.58 },
			{-14.91, -77.96, -82.53, -81.64, -81.37, -78.78 },
			{-0.01, -0.79, -42.71, -40.7, -81.47, -75.52 },
			{-0.04, -0.01, -0.03, -2.4, -0.39, -1.04 },
			{0.0, 0.0, 0.01, 0.0, -0.02, -0.7 },
		} {
			\foreach \x [count=\m] in \y {
				\def\p{1.5}
				\def\min{-88.22}
				\def\max{0.01}
				\pgfkeys{/pgf/fpu=true, /pgf/fpu/output format=fixed}
				\pgfmathparse{((\x-\min)/(\max - \min))^\p*100}\edef\j{\pgfmathresult}
				\pgfkeys{/pgf/fpu=false}
				\node[fill=green!\j!red, fill opacity=0.5, text opacity=1, minimum width=13mm, minimum height = 5mm, text=black] at (\m,-\n) {$\x$};
			}
		}
	\end{tikzpicture} \\

	&\multicolumn{2}{c}{\small\textbf{ResNet-34 on CIFAR-100}}\\
	
	\begin{tikzpicture}[yscale=0.5, font=\small]
		\foreach \y [count=\n] in {
			{$1\times 10^{1}$},    {$1\times 10^{2}$},    {$1\times 10^{3}$},    {$1\times 10^{4}$},  {$1\times 10^{5}$}, {$1\times 10^{6}$},
		} {
			\foreach \x [count=\m] in \y {
				\node[minimum height = 5mm, text=black] at (\m,-\n) {\x};
			}
		}
	\end{tikzpicture}&\begin{tikzpicture}[xscale=1.3, yscale=0.5, font=\small]
		\foreach \y [count=\n] in {
			{01.00, 34.60, 01.00, 01.00, 01.00, 01.00 },
			{01.00, 51.33, 01.00, 01.00, 01.00, 01.00 },
			{01.18, 01.00, 01.00, 01.00, 01.00, 01.00 },
			{72.41, 12.66, 01.00, 01.00, 00.92, 01.00 },
			{74.26, 73.80, 15.19, 01.05, 01.11, 01.00 },
			{73.77, 75.22, 62.94, 34.01, 18.35, 05.69 },
		} {
			\foreach \x [count=\m] in \y {
				\def\p{1}
				\def\min{1}
				\def\max{75.22}
				\pgfkeys{/pgf/fpu=true, /pgf/fpu/output format=fixed}
				\pgfmathparse{((\x-\min)/(\max - \min))^\p*100}\edef\j{\pgfmathresult}
				\pgfkeys{/pgf/fpu=false}
				\node[fill=green!\j!red, fill opacity=0.5, text opacity=1, minimum width=13mm, minimum height = 5mm, text=black] at (\m,-\n) {\x};
			}
		}
	\end{tikzpicture} & \begin{tikzpicture}[xscale=1.3, yscale=0.5, font=\small]
		\foreach \y [count=\n] in {
			{-75.11, -7.89, -77.46, -77.31, -76.95, -77.46 },
			{-72.98, 0.0, -75.23, -74.68, -74.7, -73.79 },
			{-70.32, -68.22, -64.48, -64.8, -61.12, -61.96 },
			{-0.01, -58.94, -56.6, -45.25, -47.44, -51.16 },
			{0.0, -0.3, -50.0, -58.16, -41.92, -28.91 },
			{-0.01, -0.02, -4.76, -28.26, -42.86, -34.76 },
		} {
			\foreach \x [count=\m] in \y {
				\def\p{1.5}
				\def\min{-77.46}
				\def\max{0.0}
				\pgfkeys{/pgf/fpu=true, /pgf/fpu/output format=fixed}
				\pgfmathparse{((\x-\min)/(\max - \min))^\p*100}\edef\j{\pgfmathresult}
				\pgfkeys{/pgf/fpu=false}
				\node[fill=green!\j!red, fill opacity=0.5, text opacity=1, minimum width=13mm, minimum height = 5mm, text=black] at (\m,-\n) {$\x$};
			}
		}
	\end{tikzpicture} \\
		& &\\[-3.75ex]
	\bottomrule
\end{tabular}

\end{adjustwidth}
	\end{table}

		\vspace{-6pt}

	\begin{table}[H]
	
		\caption{Top-1 accuracy for ResNet-20 on CIFAR-10, with an initial embedding of 16 feature maps, with different unstructured pruning targets, for SWD with different values of $a_{min}$ and $a_{max}$. Depending on the pruning rate, the best values to choose are not always the same. If, for each pruning target, we picked the best value among these, SWD would outclass the other technique from Table~\ref{unstructured} by a larger margin. {The best performance for each target is indicated in bold.}}
		\label{gridTradeoff}
\small

\begin{tabular}{m{1.8cm}<{\centering}m{2.5cm}<{\centering}m{2.5cm}<{\centering}m{2.5cm}<{\centering}m{2.5cm}<{\centering}}
	
	\toprule
	\multicolumn{5}{c}{\textbf{Influence of \boldmath$a_{min}$ and \boldmath$a_{max}$}}\\
	
	\multicolumn{1}{c}{\boldmath$a_{min}$} & \multicolumn{1}{c}{\textbf{0.1}} & \multicolumn{1}{c}{\textbf{0.1}} & \multicolumn{1}{c}{\textbf{1}} & \multicolumn{1}{c}{\textbf{1}} \\
	
	\multicolumn{1}{c}{\boldmath$a_{max}$} & \multicolumn{1}{c}{\boldmath{$1\times 10^{4}$}} & \multicolumn{1}{c}{\boldmath{$5\times 10^{4}$}} & \multicolumn{1}{c}{\boldmath{$1\times 10^{4}$}} & \multicolumn{1}{c}{\boldmath{$5\times 10^{4}$}} \\
	
	\midrule

	10& 
	92.38&
	92.50&
	\textbf{92.63}&
	92.56\\
	
	20& 
	92.32&
	92.57&
	92.47&
	\textbf{92.62}\\
	
	30& 
	92.53&
	92.34&
	92.45&
	\textbf{92.55}\\
	
	40& 
	\textbf{92.58}&	
	92.35&
	92.36&
	91.98\\
	
	50& 
	\textbf{92.15}&	
	92.02&
	92.08&
	92.02\\
	
	60& 
	\textbf{92.28}&
	92.09&
	92.15&
	91.89\\
	
	70& 
	\textbf{92.01}&
	91.87&
	91.69&
	91.57\\
	
	75& 
	\textbf{92.27}&
	91.89&
	90.90&
	91.70\\
	
	80& 
	\textbf{91.85}&
	91.52&
	91.37&
	91.04\\
	
	85& 
	91.44&
	\textbf{91.48}&
	90.97&
	90.7\\
	
	90& 
	\textbf{90.91}&
	90.83&
	90.15&
	90.22\\
	
	92.5& 
	\textbf{90.59}&
	90.16&
	89.88&
	89.36\\
	
	95& 
	\textbf{89.30}&
	89.00&
	88.90&
	88.28\\

	
%
%
%
%
%
%

	96& 
	88.11&
	88.64&
	\textbf{88.69}&
	87.72\\
	
	97& 
	\textbf{87.01}&
	87.0&
	86.95&
	86.67\\
	
	97.5& 
	85.76&
	86.09&
	\textbf{86.16}&
	85.91\\
	
	98& 
	83.56&
	84.27&
	\textbf{84.88}&
	85.11\\
	
	98.5& 
	75.47&
	81.62&
	\textbf{83.33}&
	82.91\\
	
	99& 
	37.24&
	74.07&
	\textbf{75.89}&
	80.2\\
	
	99.5& 
	21.61&
	47.26&
	29.35&
	\textbf{64.01}\\
	
	99.8& 
	12.27&
	11.33&
	16.47&
	\textbf{25.19}\\
	
	99.9& 
	9.78&
	\textbf{16.39}&
	11.11&
	10.9\\

	\bottomrule
	
\end{tabular}

	\end{table}

	\subsection{Experiment on Graph Convolutional Networks}

	In order to verify that SWD can be applied to tasks that are not image classification (such as CIFAR-10/100 or ImageNet ILSVRC2012), we ran experiments on a Graph Convolutional Network (GCN) based on Kipf~and~Welling\cite{kipf2016semi} on the Cora dataset~\cite{sen2008collective}. We instantiated the GCN with 16 hidden units and trained it with the Adam optimizer~\cite{kingma2014adam} with a weight decay of {$5\times 10^{-4}$} and a learning rate of {$1\times 10^{-2}$}. The dropout rate was set at 50\%.
	
\begin{figure}[H]
				\pgfplotsset{
	footnotesize,
	samples=10,
}
\begin{tikzpicture}
	\begin{scope}[scale=1.6]
		\begin{axis}[
			xlabel=Target (\%),
			xmode=log,
			log basis x={10},
			ymin=0,
			ymax=100,
			xtick={90, 10, 1, 0.1},
			xticklabels={10,90,99,99.9},
			mark size = 1pt,
			x dir=reverse,
			ylabel=Accuracy (\%),
			ylabel shift = -6 pt,
			width=0.5\linewidth,
			height=4cm,
			legend entries={Magnitude pruning, SWD},
			legend pos = south west,
			legend style = {draw=none},
			]
			
			\addplot coordinates
			{
				(90 , 83.6)
				(80 , 83.7)
				(70 , 83.8)
				(60 , 83.8)
				(50 , 83.5)
				(40 , 83.8)
				(30 , 83.7)
				(25 , 84)
				(20 , 84.3)
				(15 , 83.9)
				(10 , 83.9)
				(7.5 , 83.7)
				(5 , 83.7)
				(4 , 84.2)
				(3 , 83.4)
				(2.5 , 82.2)
				(2 , 81)
				(1.5 , 79.7)
				(1 , 74.9)
				(0.5 , 65.2)
				(0. , 14.7)
				(0.1 , 14.7)
				
			};
			\addplot coordinates
			{
				(90 , 82.9)
				(80 , 84)
				(70 , 83.9)
				(60 , 83.1)
				(50 , 83.7)
				(40 , 83.7)
				(30 , 83.3)
				(25 , 83.0)
				(20 , 83.5)
				(15 , 83.3)
				(10 , 82.7)
				(7.5 , 83.0)
				(5 , 82.6)
				(4 , 83.0)
				(3 , 83.9)
				(2.5 , 81.6)
				(2 , 81.9)
				(1.5 , 78.4)
				(1 , 77.1)
				(0.5 , 74.9)
				(0.2 , 58.9)
				(0.1 , 59.3)
			};
		\end{axis}
	\end{scope}
\end{tikzpicture}

		\caption{Magnitude pruning~\cite{han2015learning} and SWD applied on a Graph Convolutional Network\cite{kipf2016semi} on the Cora dataset~\cite{sen2008collective}.}
		\label{GCNTradeoff}
	\end{figure}
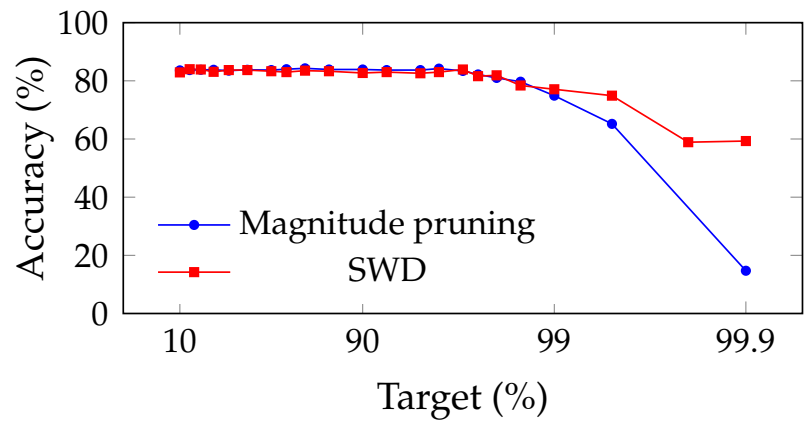

	Pruning models introduced severe instabilities when training with the original number of epochs per training, set to 200, which is why we trained models for 2000 epochs instead.
	For SWD, we set {$a_{min}=1\times 10^{-1}$} and {$a_{max}=1\times 10^{6}$}. For magnitude pruning, models were pruned across 5 iterations, with each fine-tuning lasting 200 epochs, except for the last one, which lasted 2000. The results are reported in Figure~\ref{GCNTradeoff}.
		\vspace{-6pt}

\subsection{Ablation Test: The Need for Selectivity}

	Section~\ref{sectionGridsearch} studied the sensitivity of performance to the pace at which SWD increases during training. However, we need to show the necessity of its other characteristic: its selectivity. Indeed, SWD is only applied to a subset of the network's parameters.
	
	We ran this ablation test using ResNet-20, with an initial embedding of 64 feature maps, on CIFAR-10, with unstructured pruning, and under the same conditions as stated in Section~\ref{cifarConditions}. Without any fine tuning, we compared three cases: (1) using only simple weight decay, (2) using a weight decay that grows in the same way as SWD, and (3) SWD.
	
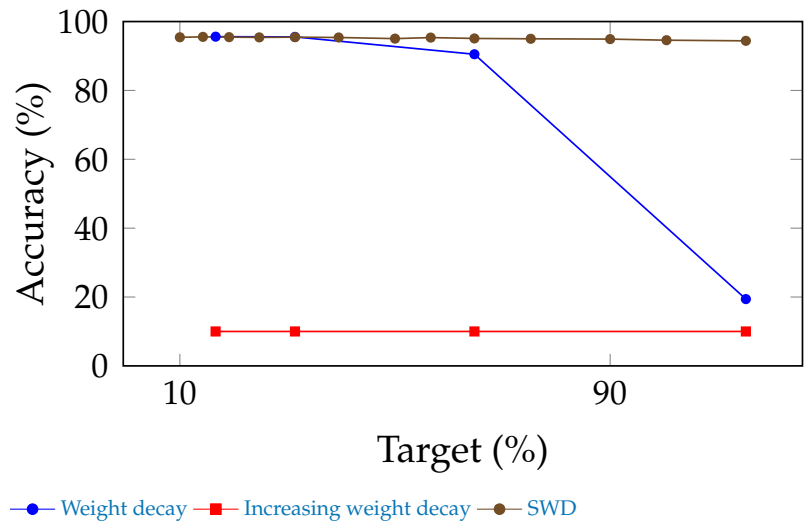
\begin{figure}[H]
				\pgfplotsset{footnotesize,samples=10}
\begin{tikzpicture}
	\begin{scope}[scale=1.6]
		\begin{axis}[
			xlabel=Target (\%),
			xmode=log,
			ymin=0,
			ymax=100,
			mark size = 1pt,
			xtick={90, 10, 1, 0.1},
			xticklabels={10,90,99,99.9},
			x dir=reverse,
			ylabel=Accuracy (\%),
			ylabel shift = -6 pt,
			width=0.50\textwidth,
			height=1.75in,
			legend entries={Weight decay, Increasing weight decay, SWD},
			legend to name=named,
			legend columns=-1,
			legend style = {draw=none},
			]
			
			\addplot coordinates
			{
				(75 , 95.61)
				(50 , 95.54)
				(20 , 90.5)
				(5 , 19.4)
			};
			\addplot coordinates
			{
				(75 , 10.00)
				(50 , 10.00)
				(20 , 10.00)
				(5 , 10.00)
				
			};
			\addplot coordinates
			{
				(90 , 95.43)
				(80 , 95.55)
				(70 , 95.47)
				(60 , 95.40)
				(50 , 95.46)
				(40 , 95.37)
				(30 , 95.04)
				(25 , 95.34)
				(20 , 95.09)
				(15 , 94.99)
				(10 , 94.90)
				(7.5 , 94.58)
				(5 , 94.4)
			};
			
		\end{axis}
	\end{scope}
\end{tikzpicture}
\linebreak \ref{named}
		\caption{Ablation test: SWD without fine-tuning is compared to a network that has been pruned without fine-tuning and to which either normal weight decay or a weight decay that increases at the same pace as SWD was applied. It appears that weight decay alone is insufficient for obtaining the performance of SWD, and that an increasing global weight decay prunes the entire network. Therefore, the selectivity, as well as the increase, of SWD is necessary to its performance.}
		\label{ablationTest}
	\end{figure}

	Figure~\ref{ablationTest} shows that neither weight decay nor increasing weight decay achieve the same performance as SWD. Indeed, the weight decay curve equates pruning a normally trained network without any fine-tuning, which is expected to be sub-optimal. Increasing global weight decay amounts to applying SWD everywhere and, thus, to pruning the whole~network.
	
	Therefore, we can deduce that (1) SWD is a more efficient removal method than the manual nullification of small weights and (2) the selectivity of SWD is necessary.
	
		{\subsection{Computational Cost of SWD}}
	
	{We measured the additional computing time caused by using SWD. {Results are presented in Table}~\ref{speedtest}. We performed experiments on ImageNet and CIFAR-10. In both cases, we obtained increased computation time of the order of 40\% to 50\%. These numbers should be put into perspective with the fact most pruning techniques come with additional epochs in training, which can easily result in doubling the computation time when compared with the corresponding baselines.}
	
		\begin{table}[H]

		\caption{{Increased 
 training duration in seconds per epoch for two different networks and datasets. Results on ImageNet ILSVRC 2012 are averaged over 5 epochs, those on CIFAR-10 are averaged over 50 epochs. Each epoch includes both training and testing.}}
		\label{speedtest}
		

\begin{tabular}{m{3cm}<{\centering}m{3cm}<{\centering}m{3cm}<{\centering}m{3.1cm}<{\centering}}
	\toprule
%
%

	\multicolumn{4}{c}{\textbf{ResNet-50 on ImageNet on 3 NVIDIA Quadro K6000}}\\
	\midrule
	SWD type & None & Unstructured & Structured \\
	Seconds per epoch & 2936 & 4352 & 4143 \\
	Increase (\%) & 0 & 48 & 41 \\
	\midrule
	\multicolumn{4}{c}{\textbf{ResNet-20 on CIFAR-10 on 1 NVIDIA GeForce RTX 2070}}\\
	\midrule
	SWD type & None & Unstructured & Structured \\
	Seconds per epoch & 55 & 77 & 78 \\
	Increase (\%) & 0 & 40 & 41 \\
	
	\bottomrule
\end{tabular}
	\end{table}
	
	\section{Discussion}
	
	The experiments on CIFAR-10 have shown that SWD  performs on par with standard methods for low pruning targets and greatly outperforms them on high ones. Our method allows for much higher targets on the same accuracy. We think that the multiple desirable properties brought by SWD over standard pruning methods are responsible for a much more efficient identification and removal of the unnecessary parts of networks. Indeed, dramatic degradations of performance, which could come from removing  a necessary parameter or filter, by error are limited by two things: (1) the continuity of SWD, which lets the other parameters compensate progressively for the loss, and (2) its ability to adapt its targeted parameters, so that the weights that are the most relevant to remove are penalized at a more appropriate time.
	
	We compared SWD to multiple methods, described in Section~\ref{spec}. Because of the large number of diverging methods in the literature, we preferred to stick to very standard ones that still serve as baselines to many works and remain relevant points of comparison~\cite{gale2019state}. The values of the hyper-parameters specific to these methods were directly extracted from their original papers. Concerning the other hyper-parameters, we ran each experiment under the same condition and initialization to separate the influence of the hyper-parameters from that of the initialization and of the actual pruning method.
	
	Because of the low granularity of filter-wise structured pruning, there is always the risk of pruning all filters of a single layer and, then, breaking  the network irremediably. This likely explains the sudden drops in performance that can be observed for reference methods in Figure~\ref{tradeoffbigfigurecifar10}. Since SWD can adapt to induce no such damage, the network does not reach random {guesses} 
	even at extreme pruning targets, such as 99.9\%.
	Our results also confirm that SWD can be applied to different datasets and networks, or even pruning structures, and yet stay ahead of the reference methods. Indeed, \mbox{Figure~\ref{GCNTradeoff}} suggests that the gains observed for a visual classification task carry over to a graph neural network trained on a non-visual task. That means that the properties of SWD are not task- or network-specific and can be transposed in various contexts (see Figure~\ref{GCNTradeoff}), which is an important issue, as shown by Gale  ~\cite{gale2019state}.
	
	\textls[-15]{Multiple observations can be drawn from the grid searches displayed in \mbox{Tables~\ref{MnistGridSearch}--\ref{structuredGridSearch}.} Experiments on MNIST show that the effect of $a_{min}$ and $a_{max}$ on both performance and post-removal drop{{ in accuracy}} 
		depends on the pruning target: the higher the target, the more dramatic the differences of behavior between given ranges of values. Upon comparison with experiments on CIFAR-10, we can tell that these behaviors are also sensitive to the models, datasets, and structures.}
	
	Both Tables~\ref{MnistGridSearch} and \ref{ExtendedGridSearch} show that high values of $a$ (or at least, of $a_{max}$ or $a_{end}$) are needed to prevent the post-removal drop in performance. This means that the penalty must be strong enough to effectively reduce weights almost to zero, so that they can be removed seamlessly. Table~\ref{ExtendedGridSearch} shows that cases of high $a_{start}$ work pretty well. This is consistent with the literature in Section~\ref{problemstatement}, which tends to demonstrate the importance of sparsity during training. However, the best results are obtained for reasonably low $a_{min}$ and high $a_{max}$, which is consistent with our arguments in favor of SWD in Section~\ref{swd}.
	
	Experiments on ResNet-20 with an initial embedding of 16 feature maps, instead of 64, revealed that these networks were much more sensitive to pruning, had a lower threshold for high $a_{max}$ values, and were more prone to local instabilities, such as the spike  for structured magnitude pruning visible in Figure~\ref{tradeOff16fm}. This is understandable: slimmer networks are expected to be more difficult to prune, since they are less likely to be over-parameterized. Because of their thinness, they also tend to be more vulnerable to layer collapse~\cite{tanaka2020pruning}, which tends to prematurely reduce the network to random guessing by pruning entire layers and, hence, irremediably breaking the network.
	
	These experiments show that the performance of SWD scales well on this slimmer network relative to the reference methods. However, the increased sensitivity to values of $a_{min}$ and $a_{max}$ highlight how sub-optimal it may be to apply the same values of these for any pruning target. Indeed, picking the best-performing combination for each target, in Table~\ref{gridTradeoff}, results in a trade-off that outmatches the reference methods by a larger margin than what Figure~\ref{tradeOff16fm} shows.
	
	However, one may notice that we generally used a different pair of values $a_{min}$ and $a_{max}$ for each experiment. Indeed, as stated previously, the behavior of SWD for certain values of $a$ is very sensitive to the overall task, and we had to choose empirically the best values we could for our hyperparameters. Moreover, for each pruning/performance trade-off figure, we used the same pair of values, while Table~\ref{MnistGridSearch} proved it to be quite sensitive to the pruning target. Therefore, it is very possible that our results are actually very sub-optimal, comparatively, to what SWD could achieve with better hyperparameter values. As finding them is very time- and energy-consuming, a method to make this process easier (or to bypass it) would be a significant improvement.
	
	Moreover, our contribution counts multiple aspects that could be expanded on and further explored, such as the penalty  and  evolution function. Indeed, we have chosen the $\mathcal{L}_2$ norm to stick to the definition of weight decay, leaving open the question of how SWD would perform with other norms. Similarly, if the exponential increase of $a$ was to bring satisfying results, other kinds of functions could be tested.
	
	{Let us add a last note about the introduced hyperparameters $a_{min}$ and $a_{max}$. Our tests suggest that a poor choice of these values may dramatically harm performance. Interestingly, however, reasonable choices ($a_{min}$ as small enough and $a_{max}$ as large enough) lead to consistently good results across datasets and architectures. These parameters have to be compared with the ones introduced by other methods. For example, in~\cite{han2015learning}, {defining multiple subtargets of pruning at various epochs during training is required,} 
		 leading to a large combinatorial search space.}
	
	Overall, the principle of SWD is flexible enough to serve as a framework for multiple variations. It could be possible to combine SWD with progressive pruning~\cite{gale2019state} or to choose gradient magnitude as a pruning criterion instead of weight magnitude.
	
	\section{Conclusions}
	
	We have proposed a new approach to prune deep neural networks continuously during training. Our theoretically motivated method, Selective Weight Decay (SWD), shows a better performance/parameters trade-off when compared with reference methods from the literature. We have shown that our method performs better while removing the need for any fine-tuning after the network is pruned.
	One great advantage of SWD is that it can  be combined with virtually any pruning criterion on any pruning structure, which opens up many possibilities. The hyperparameter $a$ and its bounds, $a_{min}$ and $a_{max}$, deserve to be studied further, leaving room for future improvements to our method.

	\authorcontributions{Conceptualization, H.T., V.G., M.L., M.A., T.H. and D.B.; methodology, H.T. and V.G.; software, H.T.; validation, H.T., V.G., M.L., M.A., T.H. and D.B.; investigation, H.T.; writing---original draft preparation, H.T.; writing---review and editing, V.G., M.L., M.A., T.H. and D.B.; visualization, H.T.; supervision, V.G., M.L., M.A., T.H. and D.B. All authors have read and agreed to the published version of the manuscript.}
	
	\funding{This work was funded by an ANRT 
 (\href{http://www.anrt.asso.fr/fr}{http://www.anrt.asso.fr/fr}{, accessed on 26 February 2022}) CIFRE scholarship between École Nationale Supérieure Mines-Télécom Atlantique Bretagne Pays de la Loire (IMT Atlantique) and Stellantis.}
	
	\institutionalreview{Not applicable.}
	
	\informedconsent{Not applicable.}
	
	\dataavailability{\url{https://github.com/HugoTessier-lab/SWD}{, accessed on 26 February 2022}).}
	
	\acknowledgments{We would like to thank GENCI, as well as Pierre Bellec, for graciously providing access to their GPUs to conduct experiments. This research was enabled in part by support provided by Calcul Québec (\url{https://www.calculquebec.ca/}{, accessed on 26 February 2022}) and Compute Canada (\url{www.computecanada.ca}{, accessed on 26 February 2022}).}
	
	\conflictsofinterest{The authors declare no conflict of interest.} 
	

\begin{adjustwidth}{-\extralength}{0cm}
\reftitle{References}

\end{adjustwidth}

\end{document}